\documentclass[pra,reprint,showkeys,nofootinbib, superscriptaddress
]{revtex4-2}
\usepackage{graphicx}% Include figure files
\usepackage{dcolumn}% Align table columns on decimal point
\usepackage{bm}% bold math
\usepackage{ragged2e}
\usepackage{amsfonts,amsmath,amssymb,amsthm}
\usepackage{booktabs}
\usepackage{enumitem}
\usepackage{hyperref}
\usepackage{float}
\usepackage{placeins}   
\usepackage[table]{xcolor}  
\usepackage{subfigure}
\usepackage[ruled,vlined,linesnumbered]{algorithm2e}

\usepackage{titlesec}
\titleformat{\section}[block]
  {\normalfont\bfseries\raggedright} % format: left/ ragged-right
  {\thesection}{1em}{}                     % label + sep
\usepackage{titlesec}
\titleformat{\subsection}[block]
  {\normalfont\bfseries\raggedright} % format: left/ ragged-right
  {\thesubsection}{1em}{}           % correct label + sep
\titleformat{\subsubsection}[block]
  {\normalfont\bfseries\raggedright} % format: left/ ragged-right
  {\thesubsubsection}{1em}{}

\newtheorem{theorem}{Theorem}
\newtheorem*{maintheorem}{Main Theorem}

\newtheorem{definition}{Definition}

\newtheorem{lemma}{Lemma}

\newtheorem{remark}{Remark}

\newcommand{\cH}{\mathcal{H}}

\newcommand{\BlackBox}{\rule{1.5ex}{1.5ex}}  % end of proof
\ifdefined\proof
    \renewenvironment{proof}{\par\noindent{\bf Proof\ }}{\hfill\BlackBox\\[2mm]}
\else
    \newenvironment{proof}{\par\noindent{\bf Proof\ }}{\hfill\BlackBox\\[2mm]}
\fi

\begin{document}
\preprint{APS/123-QED}

\title {Multi-resolution Enhancement for Full Spectrum Neural Representations}

%\title{Overcoming Low-Frequency Bias in Scientific Data Compression via Wavelet-Enhanced Implicit Neural Representations \textcolor{red}{suggestions for new title?}}% Force line breaks with \\
%\thanks{A footnote to the article title}%

% \author{%
% Yuan Ni$^{1,2}$,
% Zhantao Chen$^{1,2}$,
% Cheng Peng$^{2}$,
% Rajan Plumley$^{1,2,3}$,
% Chun Hong Yoon$^{1}$, 
% Jana B. Thayer$^{1}$,
% Joshua J. Turner$^{1,2,*}$
% }

% \affiliation{%
% $^1$Linac Coherent Light Source, SLAC National Accelerator Laboratory, Menlo Park, California 94025, USA\\
% $^2$Stanford Institute for Materials and Energy Sciences, Stanford University and SLAC National Accelerator Laboratory, Menlo Park, California 94025, USA\\
% $^3$Department of Physics, Carnegie Mellon University, Pittsburgh, PA 15213, USA
% }

\author{Yuan Ni}
\email[]{yn754@slac.stanford.edu}
\affiliation{Linac Coherent Light Source, SLAC National Accelerator Laboratory, Menlo Park, CA 94025, USA.}
\affiliation{Stanford Institute for Materials and Energy Sciences, Stanford University, Stanford, CA 94305, USA.}

\author{Zhantao Chen}
\affiliation{Linac Coherent Light Source, SLAC National Accelerator Laboratory, Menlo Park, CA 94025, USA.}
\affiliation{Stanford Institute for Materials and Energy Sciences, Stanford University, Stanford, CA 94305, USA.}
\affiliation{Walker Department of Mechanical Engineering, The University of Texas at Austin, Austin, Texas 78712, USA.}

\author{Shizhou Xu}
\affiliation{Department of Mathematics, University of California Davis, Davis, CA 95616, USA}

\author{Cheng Peng}
\affiliation{Stanford Institute for Materials and Energy Sciences, Stanford University, Stanford, CA 94305, USA.}

\author{Rajan Plumley}
\affiliation{Linac Coherent Light Source, SLAC National Accelerator Laboratory, Menlo Park, CA 94025, USA.}
\affiliation{Stanford Institute for Materials and Energy Sciences, Stanford University, Stanford, CA 94305, USA.}
\affiliation{Department of Physics, Carnegie Mellon University, Pittsburgh, PA 15213, USA.}

\author{Chun Hong Yoon}
\affiliation{Linac Coherent Light Source, SLAC National Accelerator Laboratory, Menlo Park, CA 94025, USA.}

\author{Jana B.\@ Thayer}
\affiliation{Linac Coherent Light Source, SLAC National Accelerator Laboratory, Menlo Park, CA 94025, USA.}

\author{Joshua J.\@ Turner}
\email[]{joshuat@slac.stanford.edu}
\affiliation{Linac Coherent Light Source, SLAC National Accelerator Laboratory, Menlo Park, CA 94025, USA.}
\affiliation{Stanford Institute for Materials and Energy Sciences, Stanford University, Stanford, CA 94305, USA.}

%\date{\today}% It is always \today, today,
             %  but any date may be explicitly specified

\begin{abstract}
\vspace{0.5cm}

\justifying{
\noindent
\textbf{Abstract}
Scientific data acquisition continues to outpace storage and analysis capabilities, making voxel-based representations increasingly intractable. Implicit neural representations (INRs) offer a promising solution by encoding signals through coordinate-based neural networks, serving as surrogates of data, with computational and storage requirements scaling with network complexity rather than data dimensionality. However, smaller INRs struggle to faithfully represent the multi-scale structures, high-frequency information, and fine textures that constitute a large proportion of scientific measurements. We propose WIEN-INR, a theoretically-guided hierarchical INR framework that distributes modeling across resolution scales and enables improved representation capacity through a novel enhancement network to recover subtle details. This multi-scale architecture allows smaller networks to retain the full spectrum of information while preserving training efficiency and lowering storage cost. Evaluated on distinct raw experimental measurements across scales and complexities, WIEN-INR represents a practical step toward broader adoption of neural representations in scientific workflows, delivering compact, robust, and high-fidelity representations.

\vspace{12pt}
}

\end{abstract}

% \keywords{Data compression, implicit neural representation, wavelet transformation, scattering experiments, machine learning}

\maketitle

%\tableofcontents

%================I Introduction==================
The rise and continued development of Implicit Neural Representations (INRs) have gained traction and momentum from computer vision and graphics into scientific task flows—from experimental steering \cite{chen2025implicit}, data analysis and visualization \cite{chitturi2023capturing, ni2025physicsguideddualimplicitneural,Xie2021Neural,Molaei2023Implicit}, to data compression/transmission \cite{Lu2021Compressive,Yang2023Sharing}. At the same time, as scientific facilities generate increasingly massive measurements at accelerating rates, the analysis and storage of the resulting voxel-based datasets are becoming more unsustainable, creating an urgent need for a compact and scalable data framework \cite{Thayer2024Massive,Sobolev2024}. 

Rather than treating data as discrete snapshots on voxel grids, INRs view it as samples from an underlying continuous and structured field. This perspective aligns with the observation that most scientific data does not occupy the entire ambient space, but rather lies on a lower-dimensional embedding. Hence, an INR can usually provide a more compact representation than storing values on a full coordinate grid. INRs learn a generator function that maps input coordinates (e.g., physical parameter space) to corresponding signal values (e.g., intensities), encoding the information in the parametric weights of the network \cite{essakine2025standimplicitneuralrepresentations}. It offers several distinctive advantages for scientific data pipelines, including: memory efficiency that scales with model complexity rather than resolution \cite{Lu2021Compressive,strumpler2022,Muller2022InstantNG}; it is modality agnostic, which avoids the generalization issues typical of dataset-trained deep encoding/decoding methods; it offers continuous functional fitting and differentiability with gradient access; as well as resolution independence and region-of-interest (ROI) decoding through coordinate-based queries \cite{Sitzmann2020,Mildenhall2020}. 

\begin{figure*}[p]
    \centering
    \includegraphics[width=0.95\linewidth]{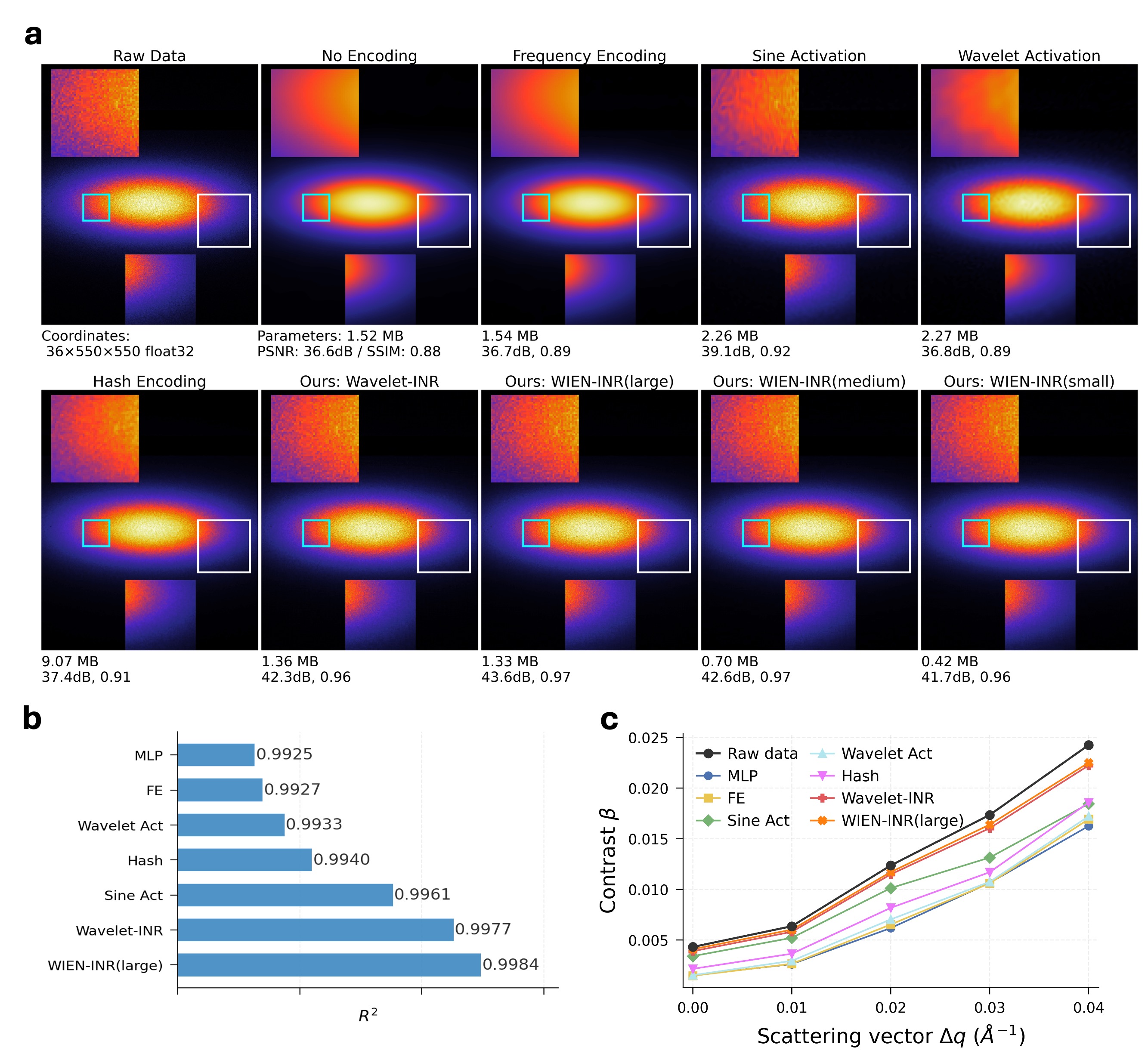}
    \caption{\textbf{Efficient implicit neural representations preserve high-frequency and physics related detail}. Comparison of implicit neural representations (INRs) of raw volumetric data (36×550×550, float32) %\textcolor{red}{what are the unit?} 
    collected from a high-resolution wide-angle coherent X-ray scattering experiment from the intermetallic alloy Cu$_3$Au \cite{ni2025cu3au}. 
    Each trained network serves as a compressed surrogate for the original volume: instead of storing the raw voxel array, the information is encoded in the network weights. \textbf{a}. A representative slice decoded by querying the trained INRs at the corresponding spatial coordinates and rendering the network outputs as pixel intensities. In all cases, the network size is smaller than the raw data representation, with parameter counts labeled beneath each subfigure. A unique challenge for coherent X-ray scattering datasets such as this is that fine-structure speckle constitutes the signal of interest, whereas in many other imaging modalities, this is treated as noise. Insets: zoomed in regions of high-speckle intensity, highlighting the superior ability of our method to capture fine details and textures. Competing INRs either over-smooth the data (No Encoding/ReLU MLP, Frequency Encoding \cite{tancik2020fourier}, Wavelet Activation\cite{Saragadam2023}, Hash Encoding\cite{Muller2022InstantNG}), produce unphysical textures (Sine activation \cite{Sitzmann2020}) or require larger networks \cite{Muller2022InstantNG}. Our method preserves the speckle and achieves the highest peak signal-to-noise ratio (PSNR) as well as a structural similarity index (SSIM) with fewer parameters counts, with metrics reported below each sub-figure. \textbf{b}. Beyond producing visually faithful representations, our method also achieves the highest R$^2$, indicating pixel-wise consistency between the decoded and the ground-truth signals. \textbf{c}. An important metric for scattering data is the speckle contrast, defined as the statistical variance of intensities divided by their mean. The contrast varies with momentum transfer $q$, i.e., a mapping that changes with distance from the peak center. We plot the $q$-dependent contrast as the region of interest is translated horizontally. Our method preserves the trend with 95\% contrast retained. Because the network is inherently compressive (with the WIEN-INR $\sim$30× smaller than the raw data), further increasing its size enables closer matching of the trends.}
    \label{fig:Cu3Au}
\end{figure*}

However, the representational capacity of conventional INRs--designed primarily for natural images, videos and scenes--is generally less reliable when modeling complex scientific data and can struggle to capture physics-relevant information. Prior scientific applications typically adopt or build upon off-the-shelf INRs as task components; our experiments show that these conventional architectures can underfit the physics-related characteristics intrinsic to scientific measurements (Fig.~\ref{fig:Cu3Au}), especially with smaller networks. Moreover, using an INR as a storage alternative requires a high-fidelity, near one-to-one representation of the underlying signal. An ideal scientific INR framework should simultaneously: (i) preserve the full spectrum of physics-related information, (ii) generalize robustly across modalities, and (iii) be computationally and storage-efficient, with fast encoding and low-latency decoding. 

These goals suggest the need for careful consideration in network design and theoretical guidance. Compact networks are favorable for fast encoding, but are often biased towards low-frequency components that tend to capture smooth components before high-frequency details \cite{Sitzmann2020,tancik2020fourier}, while larger networks offer enhanced capacity to capture finer structure, but are often costly to train and store. More broadly, identifying which features can be removed without degrading science-relevant information is non-trivial: what appears as “noise” for a given experiment may be signal to another \cite{cappello2025}. In the ideal case, increasing network size guarantees universal approximation \cite{Cybenko1989A}, meaning that sufficiently large models can represent any measurable function arbitrarily well. In practice, however, this theoretical capacity faces constraints. First, existing INRs architectures are often not parameter-efficient, with capacity not aligned to task-relevant structures. Secondly, simply enlarging the network does not always ensure convergence to the desired accuracy, particularly for fine-scale or high-frequency structures which are often more difficult to capture accurately. These limitations highlight the need for compact, full-spectrum representations that preserve both global structure and localized detail, allowing small networks to achieve high fidelity without loss of efficiency.

\begin{figure*}[tb]
    \centering
    \includegraphics[width=\textwidth]{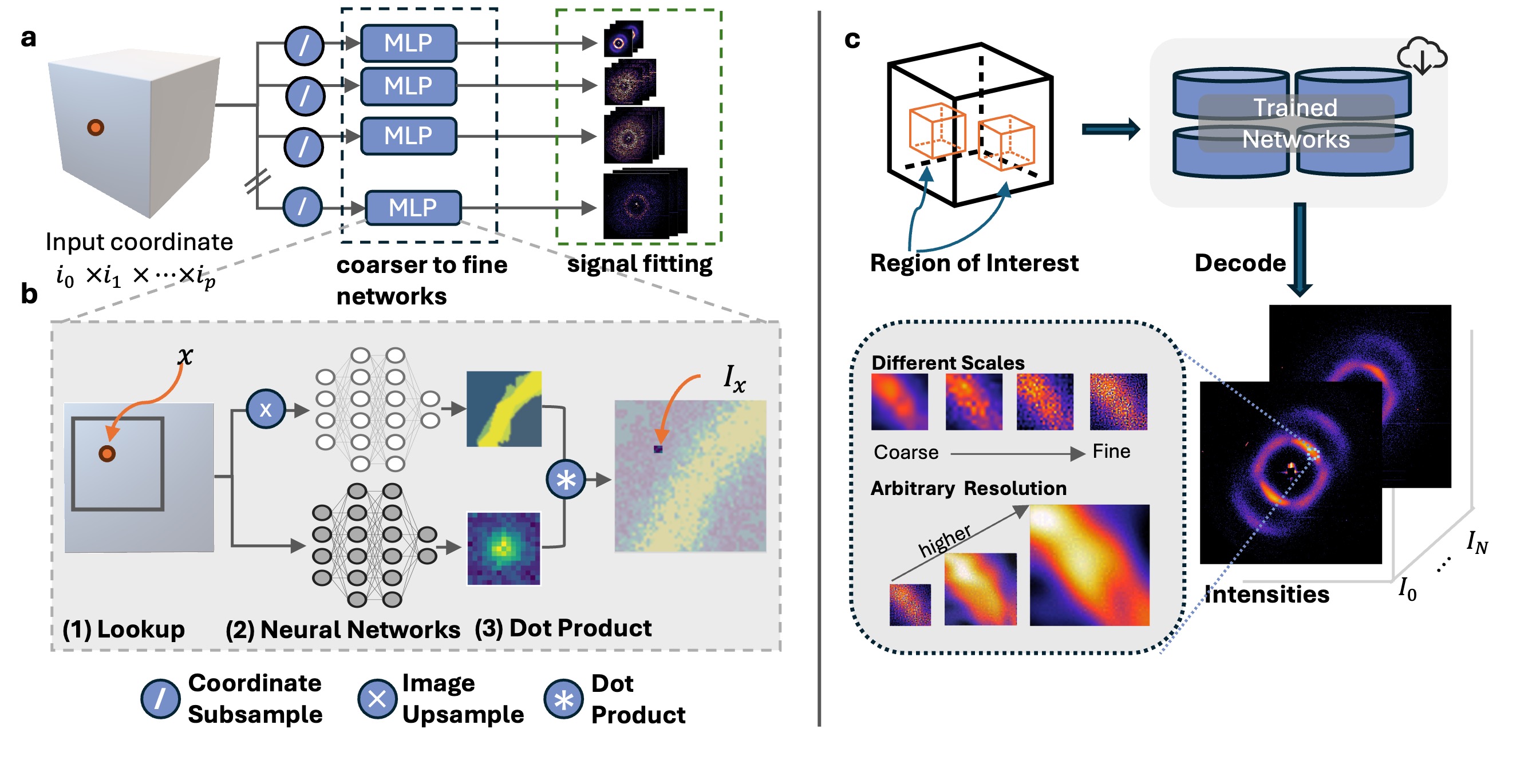}
    \caption{
    \textbf{Schematic of the WIEN-INR architecture}. \textbf{a}. We design a hierarchical encoding process. The encoding begins with the input of $p$-dimensional coordinates (e.g., detector coordinates $\mathbf{x}=[i,j]$), which are normalized and passed to separate MLPs, each trained to predict the wavelet coefficients (coarse to fine) of the signal at different resolution scales. \textbf{b}. Within each MLP, we introduce a novel enhancement module that refines the output of the previous coarser-scale MLP by: (1) extracting a local neighborhood of size $(2n_r+1)$ per input dimension, centered at the input coordinate $\mathbf{x}$ in coefficients space; (2) inputting the extracted local neighborhood to two INRs--one uses coordinates upsampled from the neighborhood to query the coarser scale MLP to provide an approximation of the higher resolution (this network is fixed), and another, a learnable INR that acts as local refinement; (3) and then applying the dot product to the outputs of these two networks to match the target finer-scale wavelet coefficients; This operation is rasterized across all coordinates of interest, refining the coarser outputs into the finer-scale outputs. \textbf{c}. After training, the decoding process allows users to query specific regions of interest (ROI) at any desired resolution using the outputs of the separate networks. To map the outputs to the true signal domain, the localized inverse discrete wavelet transform is applied to the output coefficients. Beyond the improved rate-distortion and accuracy, the hierarchical encoding and decoding process also supports multi-scale analysis (from coarse to fine detail), revealing phenomena at different frequency bands. Beyond scale, the INR-based framework supports decoding at arbitrary resolution: without interpolation, one can query increasingly fine resolutions to obtain a truly continuous representation.}
    \label{fig:arch01}
\end{figure*}
\begin{figure*}[bt]
    \centering
    \includegraphics[width = \textwidth]{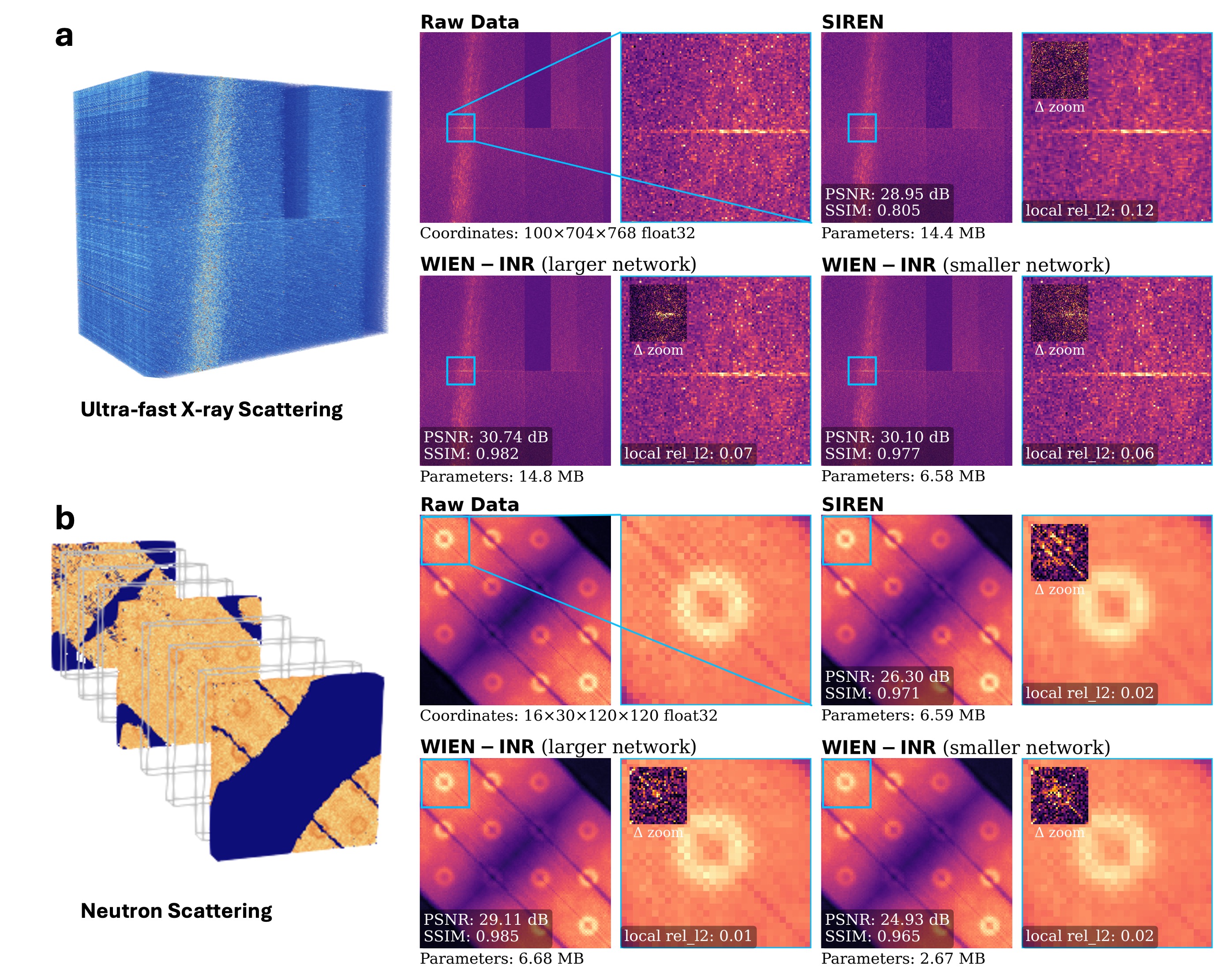}
    \caption{%\textcolor{red}{what are the units?}
    \textbf{Robustness of WIEN-INR across datasets from different physics experiments.} Results are shown for two distinct datasets collected at different facilities, one from ultra-fast X-ray scattering \cite{Chen2022ljv} at LCLS and one from 4-D neutron scattering \cite{petsch2023high} at the Spallation Neutron Source. Each volume is encoded and decoded to visualize if the network can focus training on physically important features. Representative decoded 2D slices with zoomed-in views are shown on the right. Slice-wise the peak signal-to-noise ratio (PSNR) and the structural similarity index (SSIM) metrics are reported in the inset text, while zoomed-in views also display local relative $L^2$ loss. These comparisons highlight that SIREN\cite{Sitzmann2020}—the strongest baseline— still distorts textures in the X-ray data and over-smooths the single-magnon signal in the neutron data, whereas WIEN-INR remains robust across both cases, preserving high-frequency detail and texture with fewer parameters.}
    \label{fig:detail}
\end{figure*}

\begin{figure*}[tb]
    \centering
    % \includegraphics[width=\textwidth]{new_plot/RD_01.png}%
    % \caption{Compression Ratio versus PSNR.}
    % \\
    % \includegraphics[width=\textwidth]{new_plot/RD02.png}
    \includegraphics[width=\textwidth]{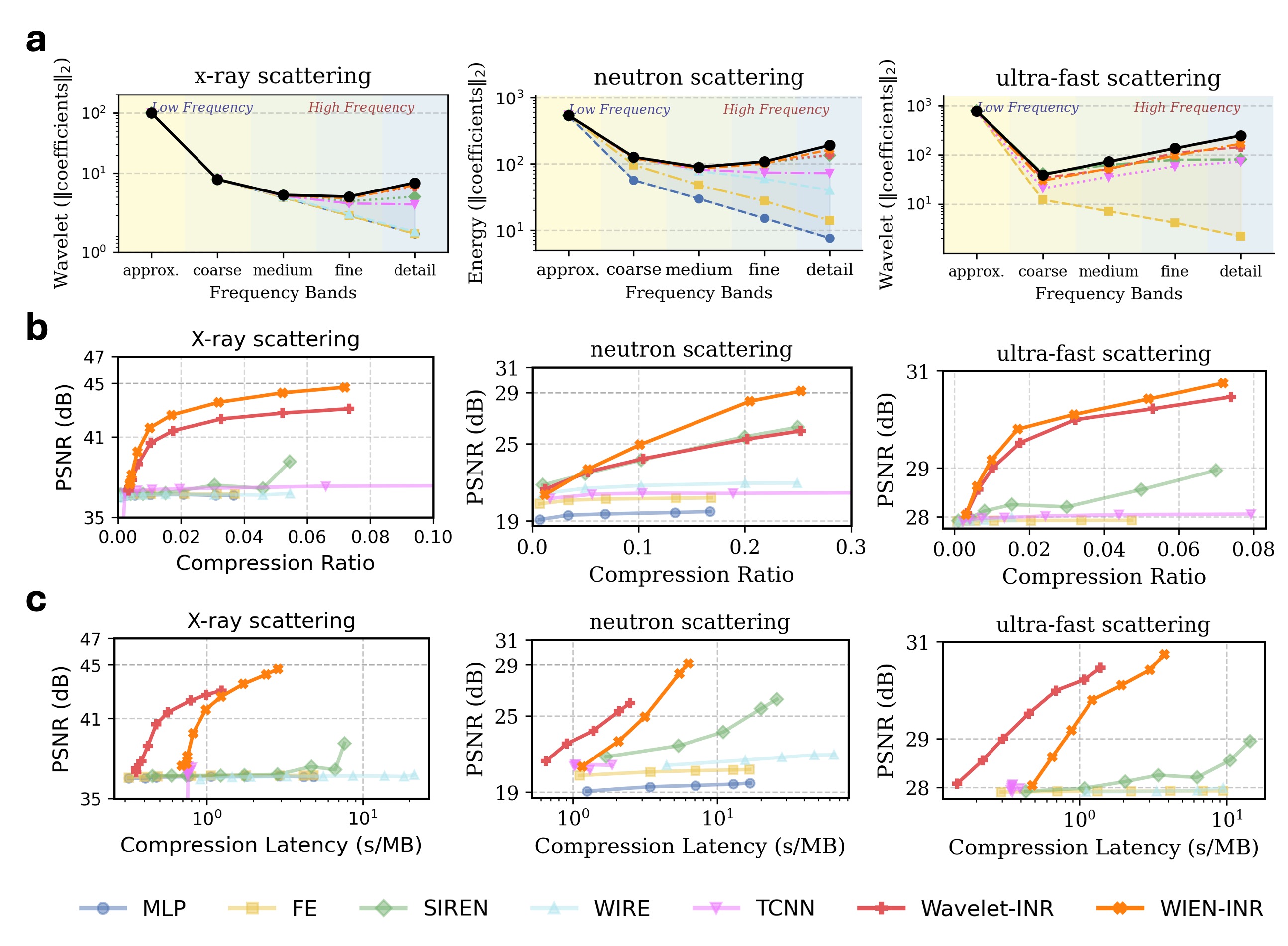}
    \caption{\textbf{Improved rate-distortion.} \textbf{a}. An important metric to quantify whether representations preserve the full spectrum of information is how well the decoded wavelet coefficients match those of the original signal. The wavelet transform is often better suited than Fourier analysis for transient or localized phenomena, as different scales capture more complete information: coarser bands represent broad, low-frequency trends, while finer bands capture localized phenomena, such as sharp transitions and high-frequency details \cite{Mallat2008}. We quantify fidelity across scales by comparing the $\ell_2$ norm of the wavelet coefficients between decoded and original signals. Results are shown across four scales and the approximation coefficients of the reconstructed discrete wavelet transform, or DWT (Haar), sub-bands for three benchmark scientific datasets. While all baselines remain close to the ground truth at coarse levels, they deviate for finer scales, WIEN-INR closely follows the true DWT distribution (black line), demonstrating robust recovery of fine-scale content and full-spectrum fidelity. \textbf{b}. Rate–distortion comparison across three scientific datasets, showing compression rate versus peak signal to noise ratio (PSNR), with stronger methods achieving higher accuracy at smaller compression rates. WIEN-INR consistently attains higher PSNR than competing INR methods at the same compression rates (i.e., with fewer network parameters). \textbf{c}. Another important aspect of the utility is encoding efficiency, i.e. the time to achieve a given accuracy level. We plot the encoding latency versus the accuracy achieved (PSNR) for the same networks used in \textbf{a} and \textbf{b}. For comparable PSNR, WIEN-INR requires less encoding time, highlighting its advantage in both fidelity and efficiency.}
    \label{fig:RD}
\end{figure*}

In this paper, we present the Wavelet Integrated Enhancement Network (\textbf{WIEN-INR}), an INR-based hierarchical framework that operates in the multiresolution wavelet domain. WIEN-INR deploys multiple subnetworks to model different time-frequency content and uses a compact enhancement module to increase the representation capacity of small coordinate-based INRs to better capture high-frequency details. The proposed architecture is supported by a mathematical theory; it is designed with the objective of retaining full-spectrum information while achieving shorter training time and smaller networks for storage. We tested WIEN-INR across a diverse set of scientific datasets, including high-resolution X-ray diffraction \cite{ni2025cu3au}, 4D inelastic neutron scattering \cite{petsch2023high}, ultra-fast X-ray scattering \cite{Chen2022ljv}, coherent diffraction imaging \cite{Favre2020} and ptychography \cite{ESRF_ID01_SiemensStar_2019}, demonstrating significant improvements in rate-distortion, spectral fidelity, texture preservation, and robustness across domains of vastly different scales and modalities. We also evaluate the sensitivity of the method to different hyperparameter choices and network configurations, demonstrating robust performance that supports generalization with minimal tuning costs.
%================II Results==================

\section*{Results}
Our approach is motivated by two key observations. First, achieving representations that faithfully preserve the scientifically relevant details while remaining compact requires architectures that can effectively capture meaningful information across the full frequency spectrum. Second, compact networks must be endowed with mechanisms that enhance their representational capacity. As illustrated in Fig.\ref{fig:Cu3Au}, existing INR frameworks, when constrained to a small parameter budget, often over-smooth fine details or introduce unphysical textures, as exemplified by a representative raw X-ray speckle dataset.
We demonstrate that WIEN-INR achieves
accurate representations that preserve fine-scale details
and textures that are physically meaningful while maintaining a compressive network size, delivering strong rate–distortion performance. 
\\

\noindent\textbf{WIEN-INR}
The design of WIEN-INR is guided by two principles. The first is that the rate–distortion can be improved by operating in a domain that better decouples and decorrelates information across scales. We achieve this by applying a wavelet transform to partition the signal into different frequency bands across scales (detailed discussion of wavelet over Fourier is provided in the Methods \ref{sec:WIEN-INR}). The second principle is that the compact coordinate-based MLPs can be endowed with additional representational power through a lightweight enhancement module, enabling them to capture subtle details without inflating the network size. Both components are modular: the wavelet transform acts as a preprocessing step, and the enhancement module can be integrated with any coordinate-based INRs.

\textit{Architecture.} Our approach addresses the limitations of using small INRs to represent the full-spectrum information, particularly high-frequency details. The wavelet transform decomposes input data into frequency bands that are better de-correlated, helping alleviate the low-frequency bias in compressive INRs by separating high-frequency details into dedicated sub-bands. We further improve model efficiency by using a shared network per scale-frequency band, taking advantage of the correlation between wavelet coefficients \cite{Mallat2008,Shapiro1993}. For the finest detail band, which is particularly challenging for small INRs to represent, we introduce a novel enhancement module that increases representational capacity without significantly increasing model size. This module infers finer-scale details conditioned on coarser-scale estimations, effectively acting as a learnable local operator that enhances the ability of the network to capture the finest details. 

\textit{Theory.} The design choices are further supported by our theoretical analysis of split INRs in the lazy regime, or the Neural Tangent Kernel (NTK) regime \citep{jacot2018ntk,lee2019wide} (formal definition is given in the Supplementary Information \ref{A:definitions}). We treat the NTK as the \emph{design variable} induced by an INR of bounded size and consider the high frequency (HF) band error. For a wavelet band indexed by $(j,i)$, let $\mathcal H_{j,i}$ denote the (fixed) band subspace and $P_{j,i}$ the associated projection operator. Let $P\in\mathbb N$ be a scalar budget equal to the total number of trainable parameters (weights and biases). Define
\[
\begin{split}
\mathcal{K}^{\mathrm{all}}(P) :=  \Big\{&K \,\Big|\,  K \text{ is the NTK of some INR that outputs} \\
& \text{all wavelet coefficients with } \leq P \text{ parameters}\Big\}.
\end{split}
\]
For feasibility of block–diagonal constructions, define the class of realizable band–restricted kernels with parameter $\theta_{j,i}$, 
\[
\begin{split}
\mathcal{K}_{j,i}(p_{j,i})\ :=\ \Big\{\,&K_{j,i}\ \Big|\  K_{j,i} \text{ is the NTK of a single head on} \\&\text{ band }  (j,i),\; |\theta_{j,i}|\le p_{j,i},\; \mathrm{Ran}(K_{j,i})\subseteq \cH_{j,i}\Big\}.
\end{split}
\]
Any band–split INR with per–band budgets $\{p_{j,i}\}$ realizes NTK $K_{\mathrm{blk}} :=\bigoplus_{(j,i)}K_{j,i}\in \mathcal{K}^{\mathrm{all}}\!\big(\sum p_{j,i}\big)$ (See Lemma~\ref{lem:additive-block} in Methods).

Let $f_0$ the initial prediction and $\mathbf{y}$ represent the true coefficients. Fix $t>0$, initial residual $r_0=f_0-\mathbf{y}$, and a high frequency error threshold $\varepsilon_{\mathsf{HF}}$. Let $P_{\textsf{sep}}(\varepsilon_{\mathsf{HF}})$ be the minimal parameters for a band–split INR to achieve HF error $\varepsilon_{\mathsf{HF}}^2$, and $P_{\textsf{mono}}(\varepsilon_{\mathsf{HF}})$ the minimal parameters among \emph{all} INRs (possibly coupled across bands):
\begin{align*}
P_{\textsf{sep}}(\varepsilon_{\mathsf{HF}})
:= \inf\Big\{ &  P\ge 0: \ \exists\ \text{block-diagonal }K\in\mathcal{K}^{\mathrm{all}}(P)\ \\ &\text{s.t. }\sum_{(j,i)\in\mathsf{HF}} \|e^{-tK}P_{j,i}r_0\|_{L^2}^2\le \varepsilon_{\mathsf{HF}}^2 \Big\},
\end{align*}
\begin{align*}
    P_{\textsf{mono}}(\varepsilon_{\mathsf{HF}}) := \inf\Big\{ & P\ge 0: \ \exists\ K\in\mathcal{K}^{\mathrm{all}}(P)\ \text{s.t. } \\ & \sum_{(j,i)\in\mathsf{HF}} \|e^{-tK}P_{j,i}r_0\|_{L^2}^2\le \varepsilon_{\mathsf{HF}}^2 \Big\}.
\end{align*}

\begin{maintheorem}[\textbf{Split is capacity–optimal at fixed HF target}]
\label{thm:sep-leq-mono}
Let $P_{\textsf{sep}}(\varepsilon_{\mathsf{HF}})$ and $P_{\textsf{mono}}(\varepsilon_{\mathsf{HF}})$ be defined as above. Under NTK linearization and Lemma~\ref{lem:additive-block}, 
\[
P_{\textsf{sep}}(\varepsilon_{\mathsf{HF}}) \;=\; P_{\mathrm{opt}}(\varepsilon_{\mathsf{HF}})
\;\;\le\;\; P_{\textsf{mono}}(\varepsilon_{\mathsf{HF}})\,.
\]
Equality holds iff the monolithic model’s realized NTK at optimum is block–diagonal over $\bigoplus_{(j,i)\in\mathsf{HF}}\cH_{j,i}$ and induces the same optimal band allocation.
\end{maintheorem}

\begin{remark}[Training-time and spectral bias]
\label{rem:time-spectral-bias}
The above result uses only parameter-budget constraints and holds for \emph{any} $t>0$. It does not exploit or assume frequency–dependent contraction rates. In particular, we do \emph{not} assume (nor rule out) inequalities such as $\lambda_{\min}(K|_{\cH_{\mathrm{LF}}}) \gg \lambda_{\min}(K|_{\cH_{\mathrm{HF}}})$ (precise definitions in Supplementary Information ~\ref{a:proof_thm_sep-leq-mono}) that encode spectral bias. Empirically, gradient descent fits low–frequency content faster than high–frequency signals (spectral bias \cite{rahaman2019spectral}), and periodic activation INRs (e.g., SIREN) mitigate \cite{sitzmann2020siren} but do not eliminate this effect as observed in our experiments. Consequently, under a small training time budget, the observed gap between split and monolithic designs can be larger than our parameter-only analytic result. A rigorous theory that jointly accounts for parameter budgets, spectral bias, and finite-time optimization is an interesting direction for future work.
\end{remark}

\begin{table*}[bt]
    \includegraphics[width=\textwidth]{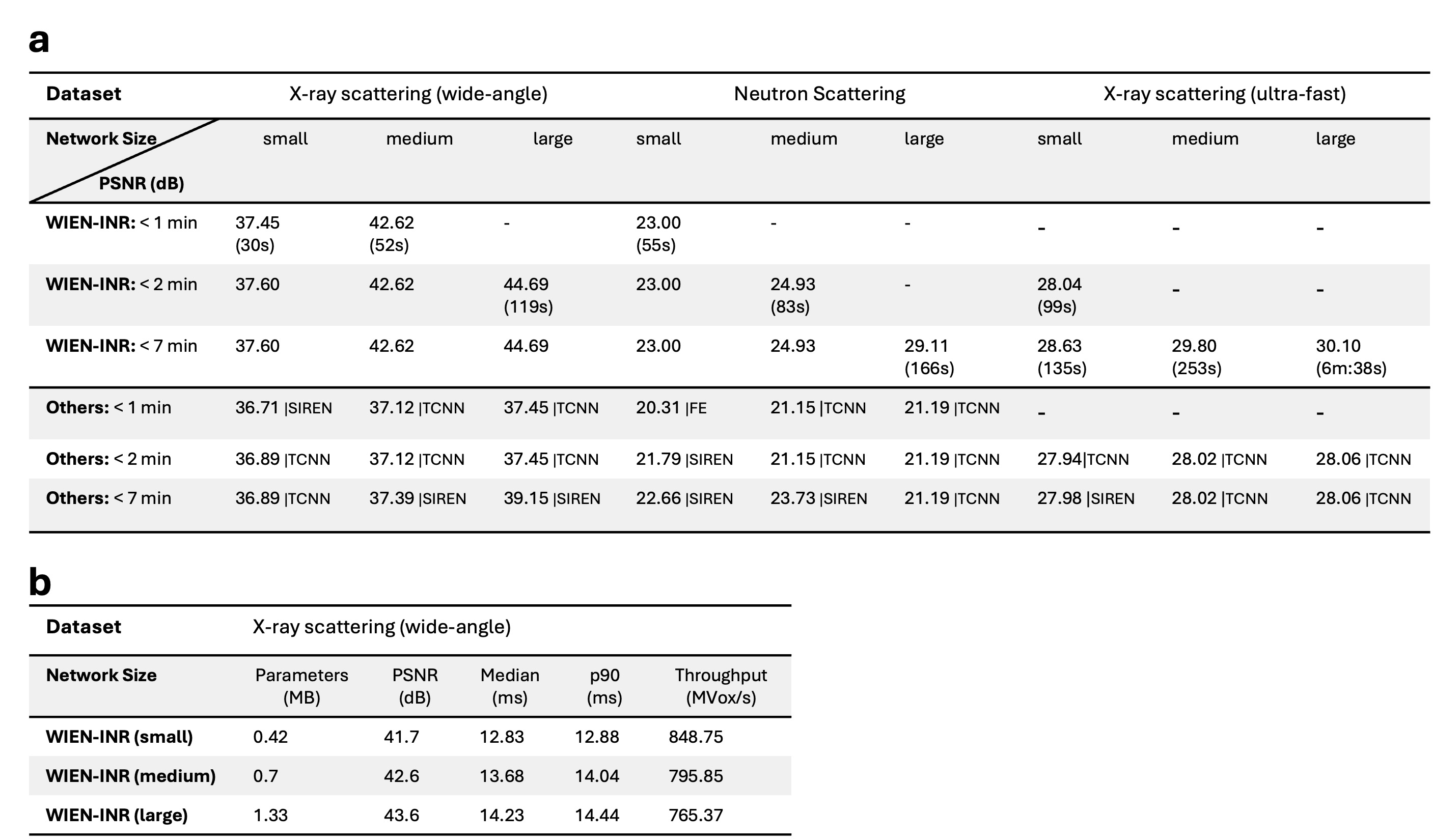}
    \caption{\textbf{Encoding and decoding latency.} Implicit neural representations (INRs) are generally slow at encoding. \textbf{a.} For comparable network sizes (small, medium, large), we evaluate accuracy after 1, 2, and 7 minutes of training. Across all three benchmark datasets, WIEN-INR consistently outperforms competing methods: for each comparable network size, it achieves the highest PSNR within the same training time. This demonstrates its utility when encoding time is critical. \textbf{b}. The decoding process only requires network evaluation at inference over the region of interest. We report the decoding time for reconstructing the full 41.5 MB volume using WIEN-INR with a small (0.42 MB), medium (0.7 MB), and large (1.33 MB) network configuration. Reported metrics include median latency (ms), 90th-percentile latency (ms), and throughput (MVox/s).}
    \label{table:speed}
\end{table*}

\begin{figure*}[tb]
    \centering
    \includegraphics[width=\linewidth]{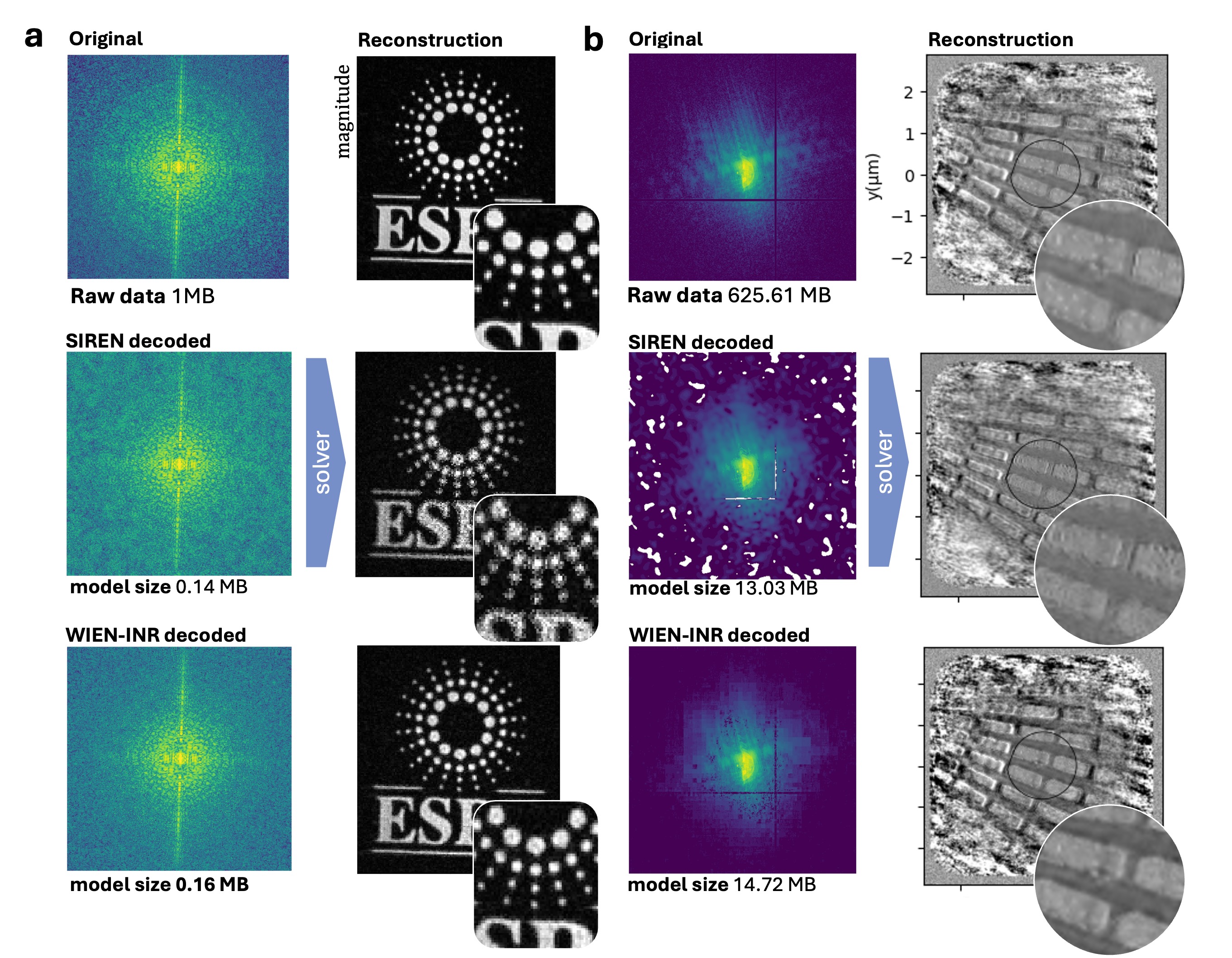}
    \caption{\textbf{Reconstruction quality using the decoded signal.} For tasks where information is revealed through subsequent reconstruction, it is crucial that the decoded signal minimally distorts reconstruction. This task is highly nontrivial because, during encoding/training, losses are typically minimized only in the measurement space, which does not directly translate into reconstruction fidelity. \textbf{a.} A use case of reconstructing an ESRF logo from INR-encoded and -decoded coherent diffraction data \cite{Favre2020}, using a combination of the error reduction (ER) and the hybrid input-output (HIO) algorithm. We plot the magnitude of the reconstruction space: the top row shows reconstruction using the original measurements without neural compression; the middle row shows reconstruction using the best-performing baseline network (SIREN 0.14 MB); and the bottom row shows reconstruction with WIEN-INR (0.16 MB). Visual degradation is evident in the SIREN case, while WIEN-INR retains higher edge contrast. \textbf{b.} Reconstruction of a ptychography measurement~\cite{cherkas2017} (625 MB) at a higher compression rate using SIREN (13.03 MB) and WIEN-INR (14.72 MB). Both networks encode and decode the diffraction images, followed by reconstruction through joint optimization of the probe and object using the alternating projection (AP) algorithm. In the central region of the reconstructed object, SIREN exhibits clear smearing, whereas WIEN-INR preserves sharp details.
}
    \label{fig:recon}
\end{figure*}
Based on the above theoretical justification, we propose a new architecture, with schematic shown in Fig.~\ref{fig:arch01}. The network is trained by minimizing the $\ell_2$ loss between the network's outputs and the wavelet coefficients at each wavelet scale (Eq.~\eqref{eq:loss}). A novel enhancement module is introduced to further improve the representation power of small networks towards high-frequency information. During training, we fix a target compression level and allocate the total parameter budget across scales according to the intrinsic data dimensionality: coarser scales are assigned smaller subnetworks, while finer scales receive larger subnetworks to better capture details.

We evaluate WIEN-INR on distinct scientific datasets, including wide-angle coherent X-ray scattering \cite{ni2025cu3au}, inelastic neutron scattering \cite{petsch2023high} and ultra-fast X-ray scattering \cite{Chen2022ljv}, each probing different physical properties. A detailed description of the datasets, including their physical origins and experimental objectives, is provided in Supplementary Information~\ref{appendix:data}. For comparison, we benchmark INRs designed with frequency awareness or multi-resolution considerations, including ReLU MLP, frequency encoding (FE) \cite{tancik2020fourier}, sinusoidal activation (SIREN) \cite{Sitzmann2020}, wavelet activation (WIRE) \cite{Saragadam2023}, multiresolution hash encoding (TCNN) \cite{Muller2022InstantNG}, and a proposed baseline approach WAVELET-INR (Eq.~\eqref{eq:wavelet_INR}). These methods address frequency representation challenges through different mechanisms: specialized input encodings, activation functions, or architectural designs. 
\\

\textbf{Improved rate-distortion}
We quantify accuracy using the peak signal-to-noise ratio (PSNR), the structural similarity index (SSIM), and pixel-wise correlation. The compression rate is defined as $$\text{Compression Rate} = \frac{P b_{\text{param}}}{N b_{\text{data}}},$$
where $P$ is the number of network parameters, $N$ the dimensionality of the original tensor, while $b_{\text{param}}$ and $b_{\text{data}}$ are the bits per parameter or data element, respectively.

Figure~\ref{fig:RD} presents our main numerical results, showing our method consistently achieves improved rate–distortion performance compared to all benchmarks. The results reveal three critical advantages: preservation of full-spectrum information (panel \textbf{a}), higher accuracy at comparable compression ratios (panel \textbf{b}), and higher accuracies with comparable encoding times (panel \textbf{c}). The enhancement module further differentiates WIEN-INR from the baseline WAVELET-INR approach by endowing networks of the same size with higher representation power. It provides better initialization and enables convergence to lower loss values, as demonstrated in Figure~\ref{fig:convergence}. 
\\

\noindent\textbf{Preservation of physics-related features}
The ability to preserve fine-scale features is particularly crucial for scientific data analysis, where subtle details often contain valuable physical information. As shown in Figure~\ref{fig:Cu3Au}, WIEN-INR accurately reproduces the speckle patterns in the X-ray scattering data of the binary alloy Cu$_3$Au, preserving both the intensity distribution (panel \textbf{b}) and contrast (panel \textbf{c}) that are critical for subsequent physical analysis—despite using a network more than 30× smaller than the raw data size. 

The qualitative results across multiple scientific datasets (Figure~\ref{fig:detail}) demonstrate WIEN-INR's modality-agnostic ability to preserve critical features while maintaining a compact representation. In these experiments, speckle constitutes the signal of interest, whereas in many other imaging fields, it is treated as noise and removed. Zoomed-in views highlight that SIREN \cite{Sitzmann2020}—the strongest baseline—over-smooths speckles in the X-ray scattering data and distorts textures in the neutron scattering data, whereas WIEN-INR remains robust across both cases.

For reconstruction tasks, we also evaluated WIEN-INR on compressing raw coherent diffraction imaging (CDI) \cite{Favre2020} and coherent X-ray imaging (CXI) measurements \cite{ESRF_ID01_SiemensStar_2019}. The quality preservation is non-trivial because $\ell_2$ metrics do not equally translate to reconstruction quality in downstream scientific applications. The compressed representations were subsequently decoded and processed/reconstructed using standard numerical inverse solvers, allowing us to visually assess reconstruction quality. As shown in Figure \ref{fig:recon}, the reconstructions from WIEN-INR maintain improved fidelity compared to those using SIREN.
\\

\noindent\textbf{Faster encoding with decoding latency}
Encode–decode efficiency is another practical bottleneck for deploying INRs in scientific workflows that require fast analysis, as INRs can be slow to encode. Splitting training across wavelet scales, WIEN-INR  yields smaller, downsampled learning problems per band ($\sim 2^{pj}$ in $p-$D at scale $j$, derived in SI \ref{appendix:wavelet}). Further acceleration is also possible through parallelization of the independently trained bands, since the loss is essentially separable. Panel \textbf{a} of Table~\ref{table:speed} reports the time to encode with WIEN-INR and the best-performing benchmark methods, across scales of network sizes that are trained for fixed durations. Notably, given equivalent training time with comparable memory requirements, WIEN-INR consistently achieves the highest accuracy (PSNR). The decoding times of WIEN-INR are reported in panel \textbf{b}. The decoding process involves the evaluation of the neural networks at relevant scales for specific regions of interest (ROIs), followed by the inverse discrete wavelet transform (IDWT).
\\

\noindent\textbf{Robustness to hyperparameters}
WIEN-INR is robust to hyperparameters, including the choice of wavelet filters (e.g., Haar, db4), number of wavelet decomposition levels $J$, and refinement window size $r$ for the enhancement module (see Fig.~\ref{fig:allJ}). A detailed discussion of the wavelet choices is provided in SI \ref{appendix:wavelet} and \ref{appendix:choice}.
% We find $J=4$ provides the best balance of frequency separation, while performance remains stable across wavelet bases, with Haar at $J=4, n_r=3$ yielding strong results across all benchmark datasets. 

We  further evaluated the robustness of WIEN-INR by varying the enhancement network's width and depth. Figure \ref{fig:wien_convergence} illustrates the training curves for the finest detail subnetwork. WIEN-INR consistently converges to lower losses faster with greater stability, compared to WAVELET-INR (which uses a SIREN network for the finest band) as shown in Figure \ref{fig:convergence_Siren}. With comparable budgets, the enhancement module achieves lower HF error (i.e., detail-band MSE), addressing a regime where standard architectures typically struggle.

\section*{Discussion}\label{sec:discussion}

Our method demonstrated reliable performance in achieving more efficient and accurate scientific data representation with a compressive network. Meanwhile, we identify several promising directions for further improvement.

\textit{Network compression and weight quantization.} In our experiments, we applied a simple post-training FP32 to FP16 quantization of network weights to reduce memory cost. More sophisticated approaches from the model compression literature could be explored in future work, including the deep compression pipeline and quantization-aware training methods that optimize for rate–distortion under low-precision arithmetic \cite{Han2015DeepCC,Yang2019,lu2021}. Beyond quantization, structured pruning and sparsification techniques may provide further reductions in model size \cite{hoefler2021sparsity}. We demonstrated a possible pipeline for post-training compression in SI~\ref{appendix:quantization}. 

\textit{Further speedup.} As the loss is separable across scales, each subnetwork can in principle be trained independently. The only exception is at the finest scale, where the enhancement network depends on the output of the network trained on the previous finest scale. This structure allows for a parallelized implementation, which would drastically reduce overall training time, with the runtime effectively dominated by the training of the next finest scale and the enhancement step. Additionally, in the context of INR-based representations, where encoding is traditionally slow, recent advances such as divide-and-conquer strategies, meta-learning, and block-wise training \cite{Han2025DCINR:, Yang2022TINC:} could further be adopted to accelerate training, making computationally efficient INR frameworks increasingly practical.

\textit{Learnable transformations.} Alternatively to hand-crafted wavelet transforms, one could learn the transform itself in a data-driven way, turning the framework into a dictionary-learning approach. Promising directions include parameterizing the lifting filters with small networks under perfect-reconstruction constraints, training shallow autoencoders to discover orthogonal wavelet bases, or making scattering/wavelet banks partially learnable to adapt to the data \cite{Coifman1993,Huang2022}. Finally, one can also explore parameterized analytic wavelet families or even architectures that bypass a fixed transform altogether. For example, Kolmogorov–Arnold Networks (KANs) \cite{liu2025kankolmogorovarnoldnetworks} are a flexible, data-adaptive multiscale transform that could bypass the need for a fixed basis while retaining interpretability.

\textit{Generative framework.} 
Since the current method is designed for faithful, compressive representation of data, another promising extension is into the generative setting. Particularly, the coarse-to-fine refinement naturally enables super-resolution by predicting fine-scale details from coarser representations, which we leave for future work. Additionally, wavelet-domain diffusion models have shown how coarse-to-fine decomposition can guide generative synthesis of 3D shapes \cite{hui2022wavelet}, while generative neural fields based on mixtures of implicit functions enable efficient sampling of new signals \cite{you2023generativeneuralfieldsmixtures}. In a similar spirit, our multi-resolution plus enhancement structure could be adapted to learn priors over coarse INR representations and enhancement kernels, enabling synthesis of high-fidelity scientific data.

\textit{Data normalization.}
In all experiments, we limited data preprocessing minimally— normalizing raw intensities to the range (0,$\infty$), and clipping away outliers — to preserve the generality of the method. For approximately uniform intensity distributions, we applied min–max normalization. For highly skewed distributions (e.g., strong peak-to-background contrast), we used a $\log(1+y)$ transform followed by max normalization. While this minimal strategy ensures broad applicability, more tailored preprocessing schemes, adapted to the statistics of specific experiments, could help training and improve representation accuracy.

\textit{Reconstruction based losses.} Beyond per-pixel/coefficient $\ell_2$ losses, one can further incorporate customized regularizations in the loss functions that enforce physics-related quantities and impose experimental constraints.%; we leave it as a future direction.

\section*{Conclusion}
We presented WIEN-INR for scientific data representation, designed to retain the full-spectrum information while achieving competitive rate-distortion performance. Our contributions are: (1) {Multi-resolution decomposition and fine scale enhancement}: By leveraging wavelet transforms and the enhancement module, the method naturally separates frequency content, enabling efficient learning of more spectrally controlled data representations that are often lost in conventional INRs. (2) Mathematical proof establishing our multi-scale architecture is optimally efficient for representing high-frequency information under the NTK scheme. (3) {Flexible architecture}: Each wavelet band is represented by an independent subnetwork, with enhancement modules applied selectively to scales where naive INRs are insufficient. This provides both scalability (wavelets can be computed in patches), parallelism (subnetworks can be trained independently), and control (users may choose how many bands to preserve). Additionally, per-pixel accuracy can be set by choosing an appropriate error bound tolerance through various loss functions such as MAE, MSE, SSIM, or perceptual losses. (4) {Data-agnostic applicability}: The framework operates directly on the wavelet coefficients of raw data with minimal pre-processing and generalizes across diverse scientific modalities, without tailoring to a specific domain. This multiresolution framework is well-suited to encoding complex data when full-spectrum information must be retained under a limited parameter budget. After encoding, the representation is a queryable, differentiable neural field that natively supports downstream tasks, e.g., ROI-based decoding, denoising, parameter fitting, registration, or downstream inference. The decoding stage is nearly instantaneous—requiring only forward neural network evaluations. By explicitly addressing the fundamental limitations of conventional INRs in representing complex scientific data, we expect WIEN-INR to broaden the utility of neural representations across data-intensive scientific domains.

\section*{Methods}
\noindent \textbf{Overview of INRs.} 
The central idea behind INR is to convert explicit information into implicit model weights through unsupervised training. Specifically, given an input coordinate $\bm{x} \in \mathbb{R}^{p}$, we are interested in learning a function that maps $\bm{x}$ to a quantity of interest $\bm{y} \in \mathbb{R}^{d}$. Such a mapping can be approximated by a neural network, typically a Multilayer Perceptron (MLP), $\varphi_\omega(\bm{x})$ parametrized by the weights $\omega$:
\[
\varphi_\omega: \Omega \subset \mathbb{R}^p \;\longrightarrow\; \mathbb{R}^d, \quad \mathbf{x} \mapsto \varphi_\omega(\mathbf{x}),
\]
where $p$ is the input dimension and $d$ is the output dimension (e.g., $d=1$ for scalar fields, $d=3$ for RGB values).

To query a multi-dimensional tensor $\bm{I} \in \mathbb{R}^{D_1 \times D_2 \times \ldots \times D_p}$, we evaluate the network on a coordinate grid. We denote by 
\[
\mathcal{I} = \{(i_1,\dots,i_p): 1 \leq i_k \leq D_k\}
\]
the full set of index tuples. Each index $(i_1,\dots,i_p) \in \mathcal{I}$ corresponds to a normalized coordinate $\mathbf{x}_{i_1,\dots,i_p} \in [-1,1]^p$ and a tensor value $\mathbf{I}[i_1,\dots,i_p] \in \mathbb{R}^d$. 

An INR is trained to minimize the discrepancy between its predictions and the ground-truth tensor values at all discrete coordinates:
\begin{equation}\label{eq:loss}
\min_\omega \sum_{(i_1,\ldots,i_p)\in \mathcal{I}} \mathcal{L}\left(\varphi_\omega(\bm{x}_{i_1,\ldots,i_p}), \bm{I}[i_1,\ldots,i_p]\right),
\end{equation}
where $\mathcal{L}$ denotes a suitable loss function. In practice, this full-sum objective is optimized over mini-batches by sampling from $\mathcal{I}$. The training objective in Eq.~\eqref{eq:loss} is conceptually straightforward, yet convergence is not guaranteed by the objective alone \cite{Sitzmann2020,serrano2024}.
\\

\noindent\textbf{Wavelet transformations and multi-resolution analysis.}
Inspired by wavelet theory, we explore a hierarchical INR design that encodes the wavelet coefficients of the data instead of working with the original space. In principle, a sufficiently expressive coordinate MLP could emulate multiscale transforms (and even surpass them) by virtue of universal approximation. In practice, however, optimization dynamics make this inefficient: standard INRs exhibit a well-documented spectral bias—they fit low frequencies first and struggle to recover high-frequency structure—so convergence on fine detail is slow or requires substantially larger models \cite{tancik2020fourier,Sitzmann2020}. 

The INR community has developed many notable techniques to mitigate the network's inherent spectral bias. A few approaches include positional encoding \cite{liu2024finer,tancik2020fourier}, which maps input coordinates to higher-dimensional features, hash encoding \cite{Muller2022InstantNG} which introduced an efficient multi-resolution grids of trainable features for compact representation, and alternative activation functions such as sinusoidal or wavelet activations as in SIREN \cite{Sitzmann2020} and WIRE \cite{Saragadam2023}. Additionally, more recent methods focus on improving model architectures to better capture fine details and a wider range of frequencies, such as INCODE, MFN and FR \cite{kazerouni2024incode,shi2024improved,fathony2021multiplicative}.

The question of identifying which features can be safely removed without affecting representation/reconstruction quality is highly non-trivial, and what seems like “noise” to a model may be signal to a researcher \cite{cappello2025}. A partial answer lies in selecting an appropriate analysis window—such as a block size or time interval that ensures meaningful patterns in the data are effectively captured \cite{Mallat2008}. This choice involves a tradeoff: smaller windows allow for detection of localized, high-frequency phenomena or ``anomalies", whereas larger windows tend to smooth over these details and are better suited for identifying broader, low-frequency trends \cite{Mallat2008}. From signal processing, we know that methods that rely on assumptions of stationarity or ergodicity over large windows (e.g., Fourier analysis) tend to obscure transient or localized phenomena, particularly when the underlying data are random or nonstationary \cite{Shapiro1993,Mallat2008}. 

Wavelet theory and multiscale analysis offer a compelling solution to this tradeoff by enabling the simultaneous examination of both local anomalies and global trends. In the wavelet domain, the coefficients provide a hierarchical description of the signal: some capture long-range correlations with narrow frequency bands, while others represent short-term variations associated with wide frequency bands. This multiscale representation inherently allows for a flexible balance between resolution in time (or space) against resolution in frequency (or momentum). An illustration of the wavelet transform coefficients and its time–frequency tiling is shown in Fig.\ref{fig:barbara}. Additional details on the technical details of wavelet transform are provided in SI~\ref{appendix:wavelet}.

This multiscale capability is particularly desirable in physics data analysis, where phenomena of interest often manifest across a range of spatial or temporal scales.
\\

\noindent\textbf{WIEN-INR}
While preparing this manuscript, Yu et al. \cite{yu2025} independently proposed CF-INR, which also directly encodes the wavelet coefficients for INR learning. Their approach uses a single-level Haar decomposition with self-evolving parameters, whereas WIEN-INR adopts a multi-level wavelet pyramid with a dedicated enhancement module designed for compact, high-fidelity data representation.
 
\paragraph{A multi-resolution preprocess.}\label{sec:WIEN-INR}
The first step in WIEN-INR is to apply a wavelet transform to the input tensor $\mathbf{I}$, decomposing it into frequency bands that are better de-correlated. The time–frequency localization property of the wavelet transform helps alleviate the low-frequency bias in compressive INRs by isolating high-frequency details into dedicated sub-bands. 
We denote the wavelet coefficients of general $p$-dimensional data with approximation coefficients denoted by $\mathbf{a}_J$, and detail coefficients of the $j$-th detail scale with direction $i$ as $\mathbf{d}^i_j, 1\leq j \leq J,\ 1 
\leq i \leq 2^p-1$. A straightforward approach to operate in the wavelet space is to apply a discrete wavelet transform (DWT) to the signal and then train separate small INRs to model each sub-band independently, and finally reconstruct the signal using the inverse discrete wavelet transform. The training process is the following,
\begin{equation}
\min_{\omega_j,\tilde\omega_J} \sum_{j=1}^{J} \sum_{i=1}^{2^p-1}
    \left\| \varphi^{i}_{\omega_j} - \mathbf{d}_j^i \right\|^2 
+ \left\| \phi_{\tilde\omega_J} - \mathbf{a}_J \right\|^2.
\end{equation}

Each neural network $\varphi^i_{\omega_j}, \phi_{\omega_J}: \Omega_j \mapsto \mathbb{R}^d$ \footnote{We write $\varphi^i_{\omega_j}, \phi_{\tilde\omega_J}$ as $\varphi^i_{j}, \phi_{J}$ from now on.}is typically parameterized as an MLP that maps spatial coordinates $\mathbf{x} \in \Omega_j$ at the appropriate resolution to the corresponding wavelet coefficient. 

However, this naive formulation suffers from the following drawbacks: (1) while wavelets decorrelate subbands in a second-order sense, the formulation ignores the strong statistical dependencies across scales and orientations, which are well-exploited in wavelet theory and zerotree coding \cite{Shapiro1993,Mallat2008}, (2) generic INR architectures struggle to converge for the finest scale (i.e., $\mathbf{d}^i_J$) due to their limited representational capacity when constrained to small network sizes. These noted shortcomings call for a more tailored architecture to handle the compression of the finest band, and ultimately, to produce improved quality encoded data.

\paragraph{Model correlation with a shared INR per scale.}

Induced by the geometric image regularity \cite{Mallat2008}, the significant coefficients in $\mathbf{d}^i_j$ tend to be aggregated along contours or in textured regions. Indeed, wavelet coefficients have a large amplitude where the signal has sharp transitions. At each scale $j$ and for each orientation $i$, a wavelet image coder can take advantage of the correlation between neighboring wavelet coefficient amplitudes, referred to as \textit{intraband correlation}. Note that taking advantage of this intrascale amplitude correlation is an important source of improvement for traditional compression techniques, such as in JPEG-2000 compression~\cite{Taubman2013}.

Taking this into account, instead of using independent INRs for each of the $2^p-1$ directions of the value $i$, we can use a single INR with an output dimension of $2^p-1$ such that it encodes information from all directions simultaneously. Specifically, we train a single $\varphi_{\omega_j}$ for the scale $j$ such that $\varphi_{\omega_i}:\mathbf{x} \rightarrow \mathbb{R}^{d(2^p-1)}.$ And the optimization problem becomes:
\begin{equation}\label{eq:wavelet_INR}
\min_{\omega_j,\tilde\omega_J} \sum_{j=1}^{J} 
    \left\| \varphi_{\omega_j} - (\mathbf{d}_j^1,\cdots,\mathbf{d}_j^{2^p-1}) \right\|^2 
+ \left\| \phi_{\tilde\omega_J} - \mathbf{a}_J \right\|^2
\end{equation}

Ideally, by adjusting the frequency bias of each $\varphi_{\omega_i}$ — for example via the frequency scaling parameter in SIREN or frequency encoding — we can align each network with the frequency range for scale $j$ it is intended to model (e.g., time-frequency tile in \ref{fig:barbara}). We call this new formulation in Equation \eqref{eq:wavelet_INR} \textbf{WAVELET-INR}, which serves as the foundation for our enhanced architecture.

\paragraph{Enhancement module.}
From our experiments, we observe that the finest detail band (approximately corresponding to frequency $\omega \in [\pi/2,\pi]$) is particularly challenging for small INRs to represent (Fig~\ref{fig:convergence_Siren}). While one possible remedy might be to further decompose this band (e.g., using wavelet packets  \cite{Coifman1993}), we argue that the difficulty does not stem primarily from frequency bias. Instead, it arises because generic INRs — especially in their compressive form — struggle to effectively capture and reproduce such fine-scale textures (SI Fig~\ref{fig:convergence}). To overcome this limitation, we introduce a tailored INR-based neural architecture that increases representational capacity for any coordinate-based INRs without incurring a significant increase in model size.

This is achieved by inferring finer-scale details conditioned on coarser-scale representations: for each fixed orientation $i$, suppose we can approximate scale $j$ with a light-weight INR $\varphi^i_j: \Omega_j \rightarrow \mathbb{R}^d$. Then for the next finer scale $(j-1)$ of the same $i$, the coefficient can be approximated by super-resolving $\varphi^i_j$ onto the next scale coordinates $\Omega_{j-1}$,
\begin{equation}
    \varphi^i_{j-1}:\Omega_{j-1} \rightarrow \mathbb{R}^d, \text{such that } \varphi^i_{j-1}(\mathbf{x}) = \varphi^i_{j}(\mathbf{x}) \text{ for } \mathbf{x}\in \Omega_j \subset \Omega_{j-1}.
\end{equation}
This upsampling operation is possible due to the continuity of the INR, and can be done recursively up to the finest detail $\Omega_{1}$. However, these upsampled networks do not contain higher frequency information beyond the scale on which they were trained. Instead, they provide an approximate prediction of the next scale, serving primarily as a geometric prior to guide the encoding of finer details — similar in spirit to zero-tree coding \cite{Shapiro1993}. 

To super-resolve the fine details, we introduce a light-weight predictor network using MLP that acts as a local learnable operator. Specifically, for a fixed orientation $i$, we train a predictor $P_{j \rightarrow j-1}^i : \Omega_{j-1}\rightarrow \mathbb{R}^{dr^p}$ to super-resolve the coarser upsampled predictions. It takes as input a coordinate $\mathbf{x} \in \Omega_{j-1}$ and outputs a kernel of size $r$ for each input dimension (i.e., $r^p$ in $p$ dimensions). Conceptually, this kernel corresponds to a local receptive field centered at $\mathbf{x}$. During training, for each coordinate $\mathbf{x}$ in $\Omega_{j-1}$, we: (1) Extract an $r^p$ patch from the upsampled coarser-scale INR prediction $\varphi^i_j$, centered at $\mathbf{x}$. (2) Apply the kernel produced by the predictor $P_{j \rightarrow j-1}^i(\mathbf{x})$ to this patch (via the inner product). (3) Match the resulting scalar prediction to the true wavelet coefficient $\mathbf{d}^i_{j-1}(\mathbf{x})$. The training objective is therefore:
\begin{equation}\label{eq:WIEN-INR}
    \min_\xi \sum\limits_{i,\mathbf{x}\in \Omega_{j-1}}\| \mathbf{d}^i_{j-1} - \left\langle \varphi^i_{j-1}(\mathcal{N}_r(\mathbf{x})), P^i_{j \rightarrow j-1,\xi}(\mathbf{x}) \right\rangle\|^2,
\end{equation}
where $\mathcal{N}_r(\mathbf{x})$ denotes the $r^p$ neighborhood of $\mathbf{x}$ in the upsampled coarse prediction. The complete formulation described above constitutes our proposed \textbf{WIEN-INR} framework.

This idea is reminiscent of a progressive refinement, where a network learns to predict the distortion between a coarse approximation and the true signal \cite{ni2025physicsguideddualimplicitneural}.  Note the enhancement module is stand-alone and can be applied to any coordinate-based INR. The enhancement module is essential to the success of high-frequency detail preservation. We also tested a more conceptually straightforward alternative enhancement approach that uses a residual network to model the difference $\{\mathbf{d}_{j-1}^i - \varphi_{j-1}^i\}_i$; however, we observed that this approach does not work better than simply using a larger network to learn $\mathbf{d}_{j-1}^i$ only. Details are provided in SI~\ref{appendix:residual}. 
\\

\noindent\textbf{HF Capacity with NTK Block Structure.}
For the theoretical results, we work on the hypercube $\Omega=[0,1]^p$ and assume the network is trained in the lazy/NTK regime \citep{jacot2018ntk,lee2019wide}, i.e., wide SIREN/MLP architectures with NTK parameterization and small updates linearize around initialization so that training dynamics are governed by a fixed Neural Tangent Kernel (NTK). In our setting, the INR does \emph{not} regress the signal $\mathbf{I}$ itself; rather, it directly regresses the \emph{wavelet coefficient fields} of $\mathbf{I}$. We note that one can derive the same conclusion for more general NTK or gradient flow setting by working with spectral bias towards low frequency signal. But we defer detailed analytic study to further work.
\\

\paragraph{Wavelet bands and HF error in coefficient space.}Fix an orthonormal, compactly supported wavelet basis $\{\psi_{j,i,k}\}$ on $L^2(\Omega)$ adapted to the boundary \citep{mallat2009wavelet}. For each scale $j\in\{1,\dots,J\}$ and orientation $i\in\{1,\dots,2^p{-}1\}$, let
\[
\cH_{j,i}\ :=\ \mathrm{span}\{\psi_{j,i,k}: k\in\mathbb{Z}^p\}\ \subset L^2(\Omega),  
P_{j,i}:L^2(\Omega)\to \cH_{j,i}
\]
be the associated band subspace and orthogonal projector. Let $\mathsf{HF}\subseteq\{(j,i)\}$ denote the set of high–frequency (HF) bands (e.g., all $j\ge J_0$). Writing $\mathbf{I}=\sum_{j,i,k} d_{j,i}(k)\,\psi_{j,i,k}$, the detail bands INR is trained to predict the \emph{detail coefficient fields} $\mathbf{y}=\{\mathbf{d}_1,\cdots,\mathbf{d}_J\} =\{d_{j,i}\}_{(j,i)}$ bandwise. Because the wavelet transform is an orthonormal isometry, squared error in coefficient space equals squared error in signal space. We therefore measure the HF error of a predictor $f$ (which outputs coefficient fields) by
\[
E_{\mathsf{HF}}(f)\ :=\ \sum_{(j,i)\in\mathsf{HF}}\ \|P_{j,i}(f-\mathbf{y})\|_{L^2}^2,
\]
where $P_{j,i}(f-\mathbf{y})$ means: reconstruct the bandwise residual via the inverse wavelet transform restricted to $\cH_{j,i}$ (equivalently, take the $L^2$ norm of the coefficient residual on band $(j,i)$).

\paragraph{NTK gradient flow for coefficient learning.}
For wide INRs and squared loss, gradient flow on parameters equals kernel gradient flow with a \emph{fixed} NTK $K$ value \citep{jacot2018ntk,lee2019wide}. For each time $t$, writing $r_t:=f_t-y$ (in coefficient space, equivalently in $L^2$ via the wavelet isometry), the dynamics satisfy
\[
\partial_t r_t\ =\ -K[r_t],\
\|P_{j,i} r_t\|_{L^2}^2\ \le\ e^{-2\lambda_{\min}(K|_{\cH_{j,i}})\,t}\ \|P_{j,i} r_0\|_{L^2}^2,
\]
where $\lambda_{\min}(K|_{\cH_{j,i}})$ is the smallest eigenvalue of $K$ restricted to $\cH_{j,i}$.

\begin{lemma}[Additivity and bandwise block structure]
\label{lem:additive-block}
Let the INR decompose as a sum of \emph{disjoint} heads $f=\sum_{(j,i)} f^{(j,i)}$, each with its own parameter block. Then the NTK is additive: $K=\sum_{(j,i)} K_{j,i}$. If, in addition, head $(j,i)$ is trained only on the band–$(j,i)$ coefficient target (equivalently, its effective output is $P_{j,i} f^{(j,i)}$), then $\mathrm{Ran}(K_{j,i})\subseteq \cH_{j,i}$ and $K=\bigoplus_{(j,i)} K_{j,i}$ is \emph{block–diagonal} over the orthogonal decomposition $\bigoplus_{(j,i)} \cH_{j,i}$.
\end{lemma}  A complete proof is provided in the Supplementary Material ~\ref{a:proof_lem1}.

\begin{theorem}[Block–diagonal NTK is HF–optimal under the budget]
\label{thm:block-opt}
Let $\mathcal{K}(P)\equiv\mathcal{K}^{\mathrm{all}}(P)$. Then
\begin{equation*}
\begin{split}
\min_{K\in\mathcal{K}(P)} \sum_{(j,i)\in\mathsf{HF}}\|e^{-t K} P_{j,i} r_0\|_{L^2}^2
 \;&=\; \\
 \min_{\{K_{j,i}\in\mathcal{K}_{j,i}(p_{j,i}), \,\sum p_{j,i}\le P\}}
& \sum_{(j,i)\in\mathsf{HF}}\|e^{-t K_{j,i}} P_{j,i} r_0\|_{L^2}^2 .
\end{split}
\end{equation*}
In particular, some optimizer is block–diagonal across bands: $K^\star=\bigoplus_{(j,i)\in\mathsf{HF}} K_{j,i}^\star$.
\end{theorem}
A complete proof is provided in the Supplementary Material ~\ref{a:proof_thm_block-opt}.
\\

\noindent\textbf{Implementation Details}
%\textit{Architecture.} 
For the subnetworks that represent each scale (i.e., \{$\varphi_{\omega_j},\phi_{\tilde \omega_J}$\}), we use independent SIREN networks with varying width and depth, where the sine layer frequencies are varied across scales/but fixed across experiments, which controls the distribution of frequencies the network represents. The subnetwork size is controlled solely by adjusting the hidden layer dimensions. The enhancement kernel network is also implemented using a SIREN, with tunable hidden layer width but fixed frequency parameters $\omega_0=30$ for the first layer and $\omega=30$ for the hidden layers. Details about the configurations for the Cu$_3$Au sample experiments are summarized in Table~\ref{tab:standard_methods} and ~\ref{tab:wavelet_methods}. SIREN is used as a backbone as it consistently achieved the best standalone performance across the scientific datasets examined in this work. We choose to apply the enhancement structure to the finest detail band (i.e., $\mathbf{d}_2$ level to the $\mathbf{d}_1$ level). This choice is data-dependent (e.g., distribution of information across wavelet bands), but we found it sufficient for all evaluated scientific datasets, where the most challenging information to compress resides in the finest details. In practice, the enhancement network can be applied at any scale.
\\
We implemented our framework on one NVIDIA A100 GPUs with 40 GB memory. To optimize inference and decompression, all trained model weights are stored in half precision (2 bytes per entry). For training stability, we maintain a master copy of the parameters in full precision and use mixed-precision updates. A more detailed description of the network, training and hardware can be found in SI~\ref{appendix:training}.

\begin{acknowledgments}
This work was primarily supported by the Department of Energy, Laboratory Directed Research and Development program at SLAC National Accelerator Laboratory, under contract DE-AC02-76SF00515. C.P. was supported by the Department of Energy, Office of Science, Basic Energy Sciences, Materials Sciences, and Engineering Division. This research used computational resources of the National Energy Research Scientific Computing Center (NERSC), a US Department of Energy Office of Science User Facility located at Lawrence Berkeley National Laboratory operated under contract DE-AC02-05CH11231, using NERSC award BES-ERCAP0026843.

The authors thank Alexander N. D. Petsch for helpful discussions.

During the preparation of this work, the authors used the large language model ChatGPT by OpenAI in order to refine the language and enhance the readability of this paper. After using this tool, the authors reviewed and edited the content as needed and take full responsibility for all content in this publication.

\end{acknowledgments}

%%%%%%%%%%%%%%%%%%%% REFERENCES %%%%%%%%%%%%%%%%%%
% The best way to enter references is to use BibTeX:

% \bibliographystyle{mnras}
%\bibliographystyle{unsrt}
\bibliographystyle{unsrtnat}
\bibliography{reference} 

\clearpage 
\onecolumngrid

\section*{Extended Data Figures}
\setcounter{figure}{0}
\renewcommand{\thefigure}{S\arabic{figure}}
% (optional, fixes hyperref anchors)
\makeatletter
\renewcommand{\theHfigure}{S\arabic{figure}}
\makeatletter

\begin{figure}[htb!]
    \centering
    \includegraphics[width=0.7\linewidth]{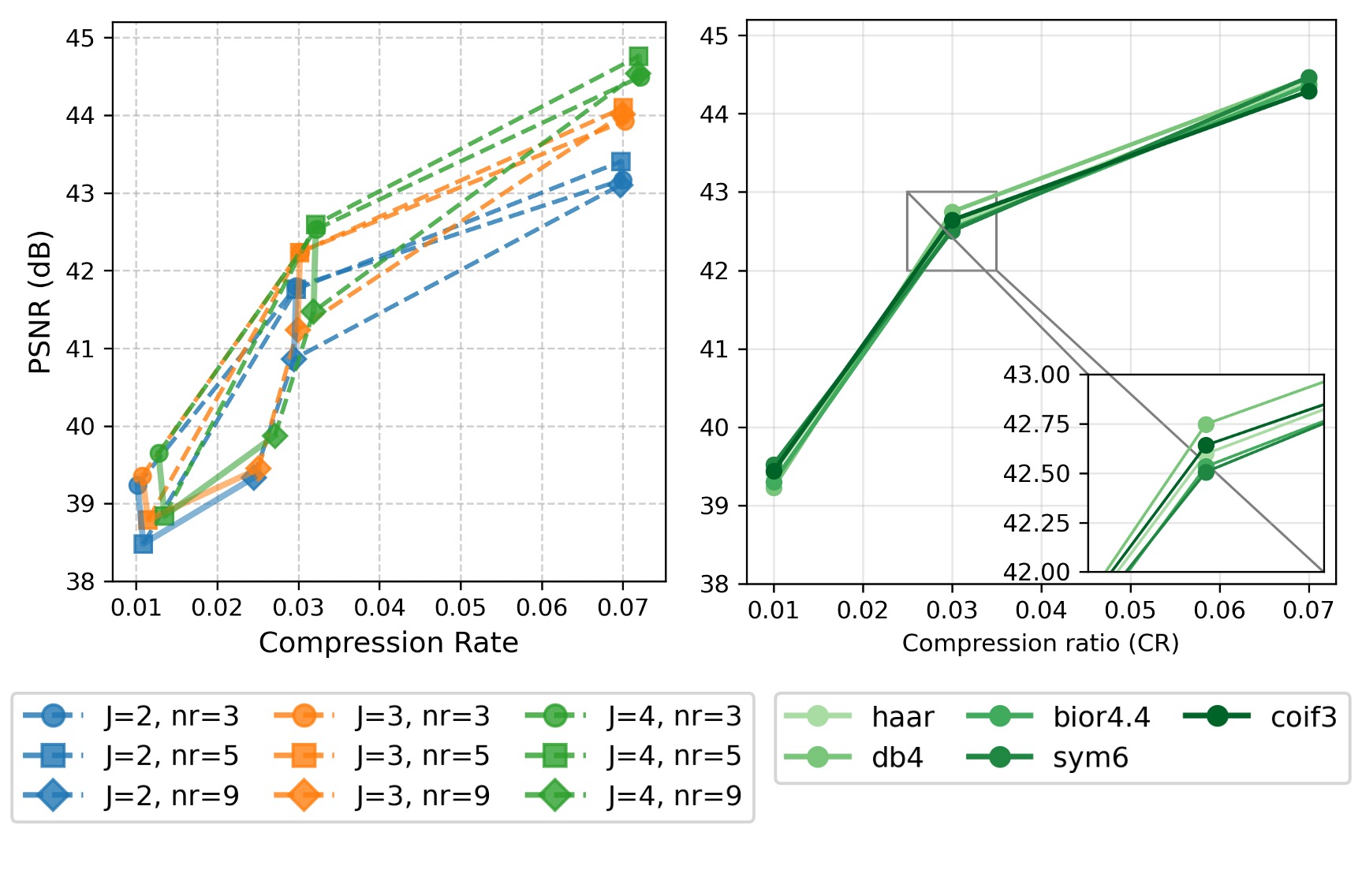}
    \caption{\textbf{Robustness to wavelet choices.} (Left) Use a fixed wavelet type with varying depth $J$ and $r$ to encode the Cu$_3$Au sample \cite{ni2025cu3au} using WIEN-INR. We plot the rate--distortion curves using Haar wavelet with different multi-resolution depths $J \in \{2,3,4\}$ and neighborhood sizes nr$ = r \in \{3,5,9\}$. We plot the compression rate (network size over raw size) versus the achieved accuracy (PSNR (dB)). Larger $J$ and $nr$ induce larger network, but potentially higher representation power. Colors indicate different $J$, markers indicate different $r$; curves with the same $r$ are connected by dashed lines, and curves with the same $J$ by solid lines. (Right) Study of using a fixed depth $J$ and $r$ but varying wavelet types. Rate-distortion curves for $J=4$ and $r=3$ are plotted, where our network is trained to encode DWT coefficients computed with five different wavelet bases: Haar, Biorthogonal 4.4 (bior4.4), Coiflet 3 (coif3), Daubechies 4 (db4), and Symlet 6 (sym6), indicated by different shades of green colors. We find $J=4$ provides the best balance of frequency separation, while performance remains stable across wavelet choices, with Haar at $J=4, n_r=3$ yielding strong results across all benchmark datasets. }

    \label{fig:allJ}
\end{figure}

\begin{figure*}[tb]
    \centering
    \includegraphics[width=0.75\linewidth]{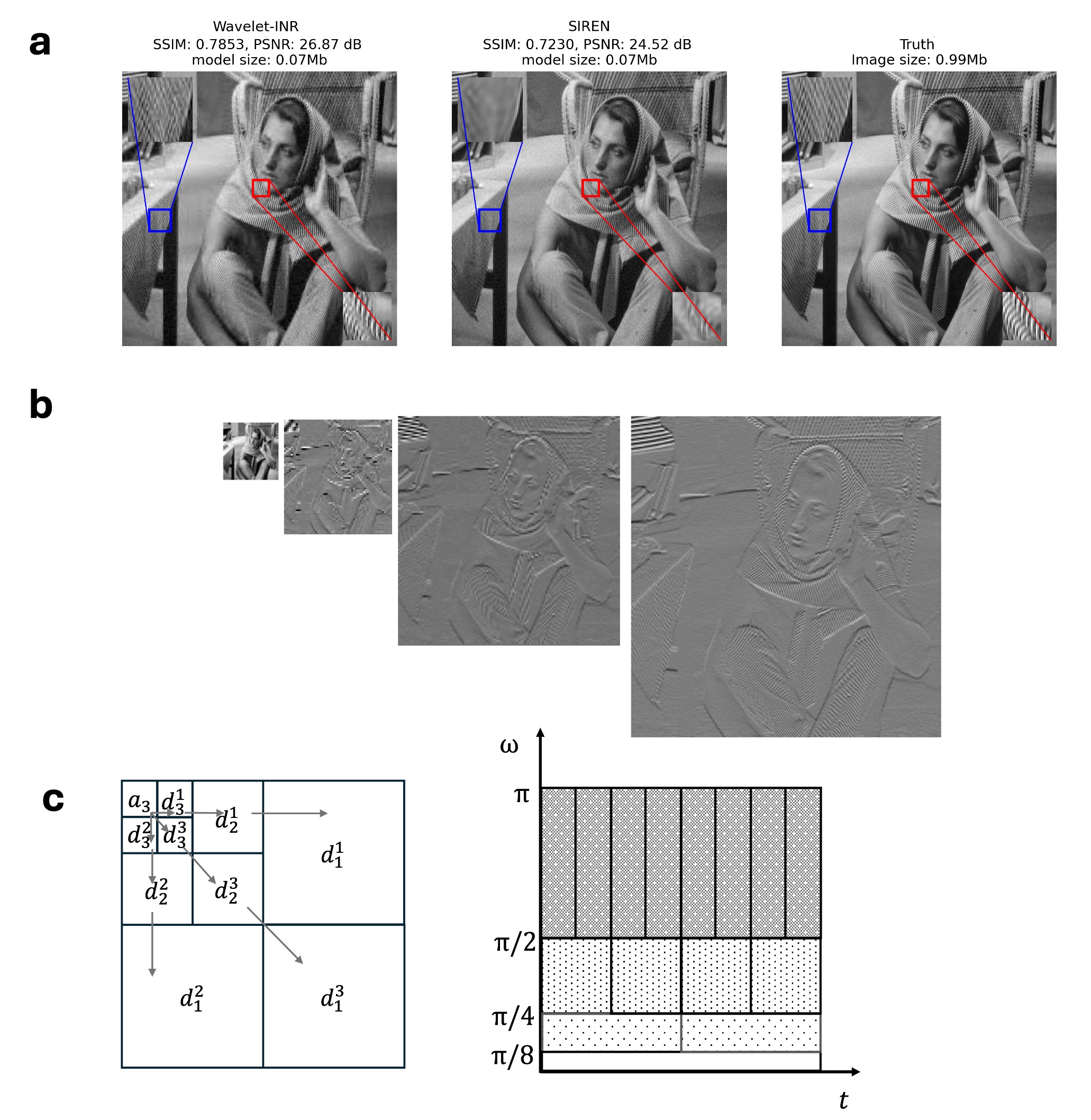}
    \caption{\textbf{Multi-resolution wavelet decomposition with implicit neural representations.} 
    \textbf{a.} Neural representation of the Barbara image using different INR approaches. Left: Separate INRs for each DWT sub-band, which is the same as our baseline WAVELET-INR. Center: A single network representing all frequency bands jointly. Right: Ground truth image. With the same total number of network parameters, the left decomposition assigns more representational capacity to high-frequency bands, yielding a representation with enhanced fine details as seen in the enlarged textures. \textbf{b}, Example wavelet coefficients of the Barbara image, showing approximation coefficients $\mathbf{a}_3$ and detail coefficients $\mathbf{d}_3^1$, $\mathbf{d}_2^1$, and $\mathbf{d}_1^1$ (from left to right) using the haar wavelet. \textbf{c,left}, Schematic of a two-dimensional wavelet transform, where the lowest-frequency approximation sub-band ($\mathbf{a}_3$) occupies the top-left quadrant, and progressively higher-frequency detail sub-bands appear toward the bottom right. Within each scale $j$, the three detail sub-bands ($\mathbf{d}_j^1,\mathbf{d}_j^2,\mathbf{d}_j^3$) capture horizontal, vertical, and diagonal orientations, respectively. \textbf{c,right}, Corresponding tiling of the time–frequency plane. Each successive scale approximately halves the bandwidth in the frequency domain, with scale $j$ responding primarily to the octave $[\pi/2^j,\pi/2^{j-1}]$, subject to overlap due to the non-ideal frequency localization of practical wavelets.
}
    \label{fig:barbara}
\end{figure*}

\clearpage
\begin{figure*}[tb]
    \centering
    \includegraphics[width=0.7\linewidth]{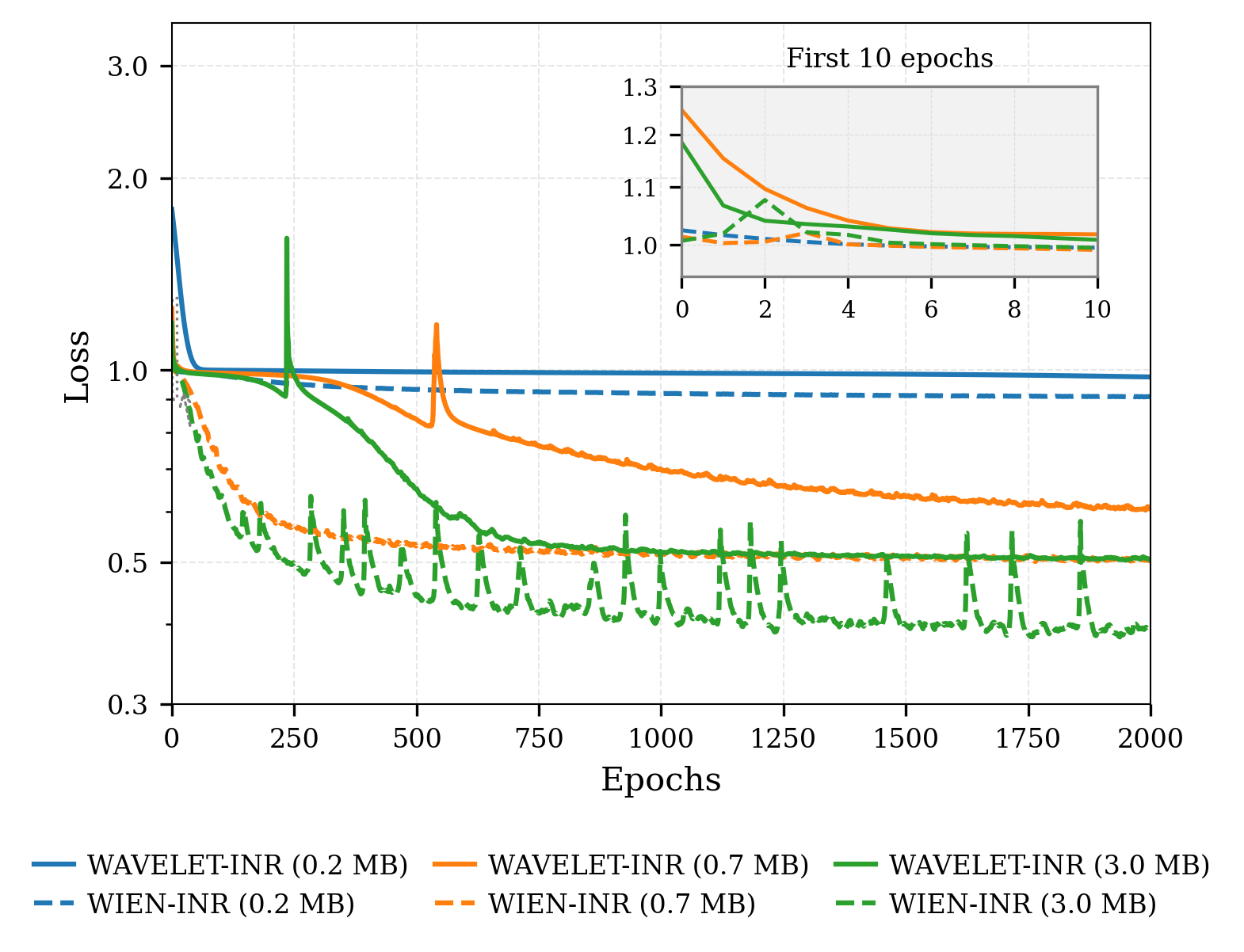}
    \caption{\textbf{Convergence of the finest scale for the Cu$_3$Au dataset \cite{ni2025cu3au}.} Training with and without the proposed enhancement on the finest detail band for the Cu$_3$Au dataset. Note that with comparable network sizes (indicated by colors), WIEN-INR (solid lines) consistently achieves faster convergence to lower final losses compared to the baseline Wavelet-INR (dashed lines). The inset plot shows the losses for the first 10 epochs, we observe WIEN-INR begin with substantially lower initial loss—a direct benefit of conditioning on coarser-level predictions rather than starting from random initialization. This demonstrates how the proposed coarser-to-fine enhancement provides both better initialization and superior final convergence for high-frequency detail representation.
}
    \label{fig:convergence}
\end{figure*}

% Training used weight decay $1e-05$ with 3-layer networks of various hidden layers. 
\clearpage

% \linenumbers
% \resetlinenumber[1] 
\section*{Supplementary Information}
\pagenumbering{arabic}\setcounter{page}{1} % restart pages
\appendix
\tableofcontents
\clearpage
\section{Data}\label{appendix:data}
%Describe the data and what physics they study. And why from the signal processing perspective they are different. \\
%message:repetition rate, speckle, different physics, challenges....
\begin{itemize}
    \item Cu$_3$Au X-ray data: The data was collected at Beamline 11-ID at the National Synchrotron Light Source II at Brookhaven National Laboratory using a Dectris Eiger1M detector with 75 micron pixel size at distance of 1.5~meters from the X-ray interaction point.  The time-series data was collected using 1-second exposures from a single-crystal Cu$_3$Au sample aligned to the (001) and (002) crystal reflections at an X-ray energy of 12.8~keV.  The temperature of the sample was maintained at 100 Kelvin using a closed-cycle helium cryostat.

    %repetition rate, speckle, what it studies, facility.
    \item 4-D inelastic neutron scattering: The data was collected on the SEQUOIA spectrometer at the Spallation Neutron
Source at the Oak Ridge National Laboratory. The experimental setup is shown in Ref.~\cite{chen2025implicit}. The sample is free to rotate by an angle about the vertical axis in the laboratory frame. A detector bank placed downstream of the incident neutron beam records the scattered neutrons with broad coverage of the four-dimensional ($\mathbf{Q}$,$\omega$) space as the sample is rotated. The crystal has the conventional high-temperature tetragonal unit cell, with $a=b\approx 3.89$,\text{\AA} and $c\approx 12.55$,\text{\AA}, and express reciprocal-space vectors as $\mathbf{Q}=H\mathbf{a}^*+K\mathbf{b}^*+L\mathbf{c}^*\equiv(H,K,L)$ in reciprocal-lattice units (r.l.u.). More details on data collection and processing are provided in Refs.~\cite{ni2025physicsguideddualimplicitneural,chen2025implicit,chitturi2023capturing,petsch2023high} %4d coordinate, q omega dependence.
    \item Ultra-fast X-ray scattering: Coherent X-ray speckle patterns were recorded from a single-crystal NiPS$_3$ sample. X-ray photon counts on an EPIX100 detector were integrated over $N_{\mathrm{frames}}$ exposures. The effects of averaging are evident when comparing a short acquisition of 100 frames ($\approx 1$ s at a 120 Hz repetition rate, e.g., at LCLS) with a longer acquisition of 34,081 frames ($\approx 5$ min). More details on data collection and processing are provided in Ref.~\cite{Chen2022ljv} %ultra-fast xray: cite hongwei paper, 120HZ which will be significantly increase to $10^{6}$\,HZ, corresponds to xxGB/sec \cite{Thayer2024Massive}. 
\end{itemize}

\section{Network and Training}\label{appendix:training}
\subsection{GPU utilization}
All experiments are trained on a single NVIDIA A100 GPU with 40 GB memory. To optimize inference and decompression, all trained model weights are stored in half precision (2 bytes per entry). For training stability, we maintain a master copy of the parameters in full precision and use mixed-precision updates. For the Cu$_3$Au sample, we report results for three model sizes—small (0.42 MB), medium (0.70 MB), and large (1.33 MB)—with peak GPU memory usage of 2.4 GB, 2.86 GB, and 3.2 GB, respectively. During training, we use a batch size of 302,500.

\newpage

\subsection{Enhancement via residual networks}\label{appendix:residual}
For the refinement module, we also experimented with a dual-INR design, where two subnetworks were trained jointly in a residual-like manner. 
Algorithmically, the setup is identical to the WIEN-INR encode scheme, except that the kernel predictor is replaced by a second INR to learn the residual part. To predit $j$-th scale from $(j+1)$ scale:  
\begin{enumerate}[label=(\arabic*)]
  \item Initiate $\mathcal{\varphi}_{\text{res}}$ with configuration $\texttt{config}_{\text{res}}$. Build coordinates for the $j$-the band as $\mathbf{x}_j$. The network trained for the coarser band is passed in as $\varphi_{\text{pre}}$.  
  \item Compute the residual: $\Delta Y_j \gets \mathbf{d}_j - \varphi_{\text{pre}}(\mathbf{x}_j)$.  
  \item Train $\varphi_{\text{res}}$ on $\mathbf{x}_j$ to predict $\Delta Y_j$, using $\ell_2$ loss.  
  \item Reconstruction is then given by $\hat{\mathbf{d}}_j \gets \varphi_{\text{pre}}(\mathbf{x}_j) + \varphi_{\text{res}}(\mathbf{x}_j)$.  
\end{enumerate}

\begin{figure}[hbt]
    \centering
    \includegraphics[width=\linewidth]{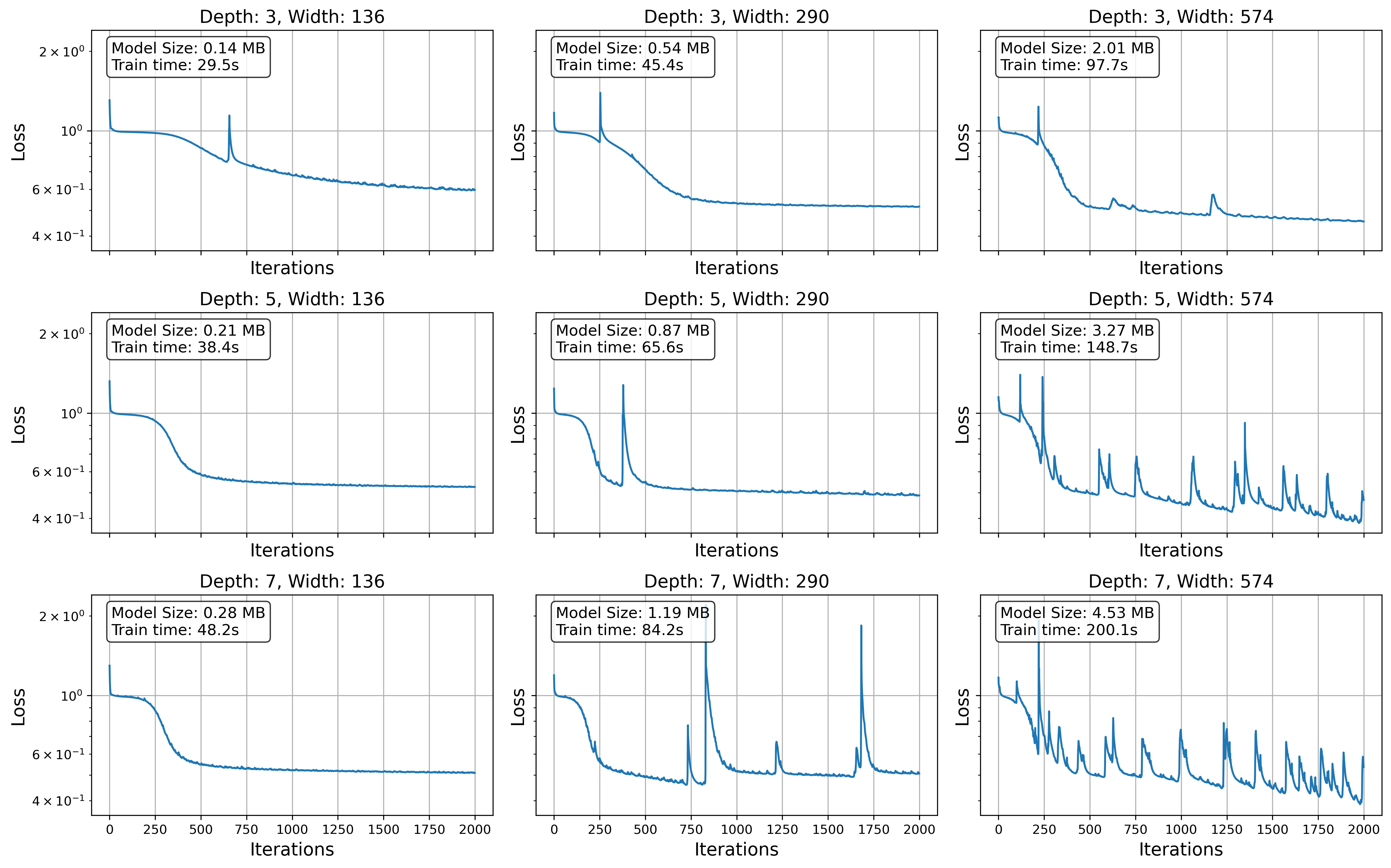}
    \caption{
    \textbf{WAVELET-INR Convergence on Cu$_3$Au sample.} Study of the convergence behavior for the finest detail band using SIREN with varying depth and width as indicated in the subplot titles. Training used a learning rate of 1e-4 and weight decay of 1e-5. These curves represent the convergence behavior for the sub-model of WAVELET-INR architecture that learns the finest detail band. While convergence improves with increasing network size, the loss never approaches zero, indicating inherent limitations in representing high-frequency details with a standard SIREN.}
    \label{fig:convergence_Siren}
\end{figure}

\begin{figure}
    \centering
    \includegraphics[width=\linewidth]{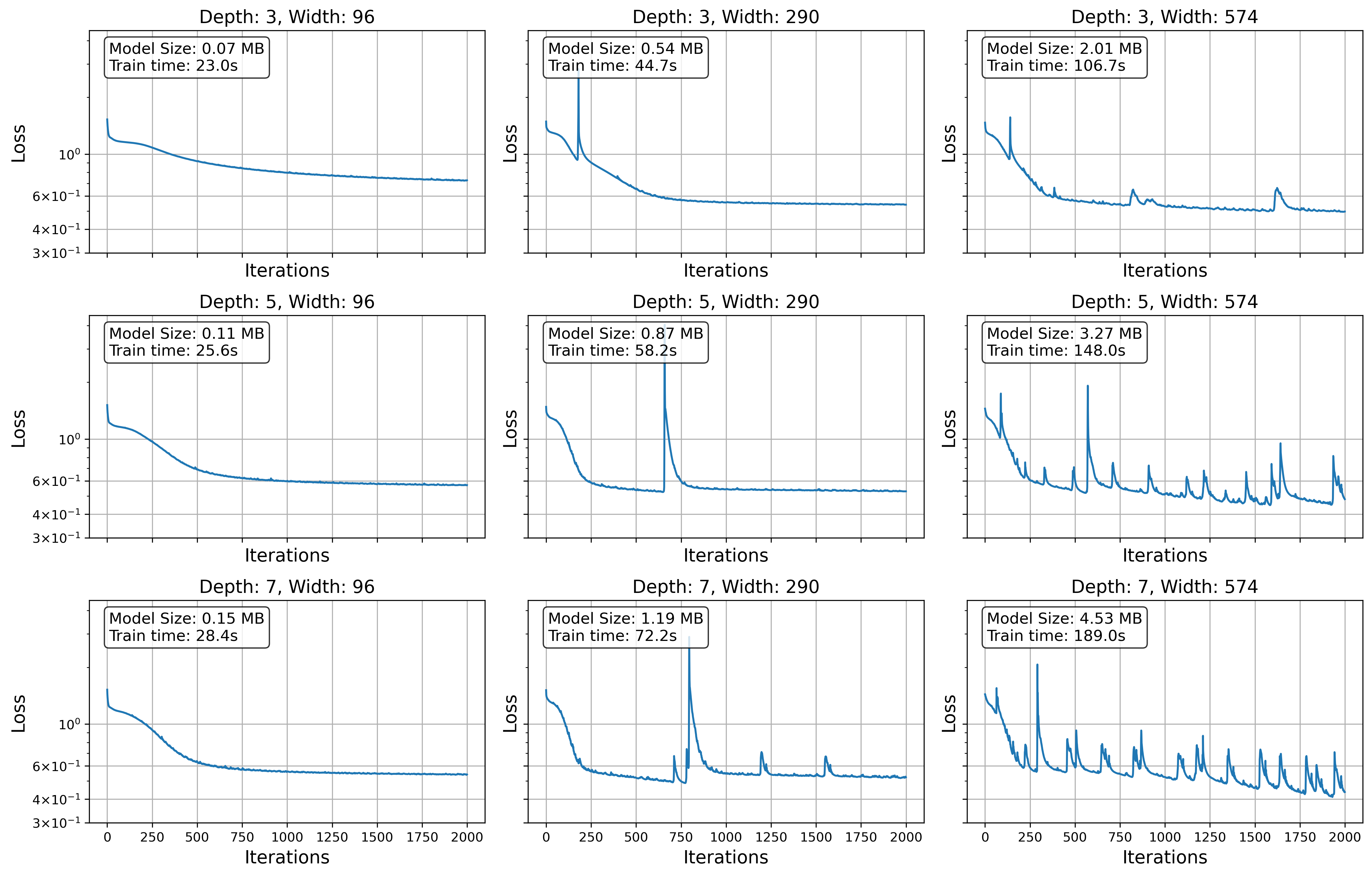}
    \caption{\textbf{ResNet convergence with dual SIREN on Cu$_3$Au sample.} We implemented a residual approach using a SIREN with varying depth and width to model only the residual component when predicting the finest detail band (learning rate = 1e-4, weight decay = 0). This approach performs no better than the results shown in the previous figure, which simply uses a SIREN to directly learn the finest band, suggesting that residual network architecture alone do not address the fundamental challenges in high-frequency representation.}
    \label{fig:res}
\end{figure}

Specifically, we tested using two SIRENs (one for $\varphi_\text{pre}$ and one for $\varphi_\text{res}$) and a mixed design (SIREN + WIRE) as inspired by \cite{Roddenberry2023,Yang2019}. However, these variations (Fig.\ref{fig:res}) did not yield consistent improvements over the single-INR baseline (Fig.\ref{fig:convergence_Siren}). The two plots compare the convergence behavior of the finest detail band when using only a SIREN network (Fig.\ref{fig:convergence_Siren}) versus using a residual network built upon the previous band (Fig.\ref{fig:res}). We explored various network configurations by varying width and depth parameters, finding no consistent advantage in the residual network approach.

\subsection{Post-training compression}
\label{appendix:quantization}
An alternative way for compressive INR begins with training a large or over-parametrized network, followed by post-processing compression techniques. This method is particularly valuable when prioritizing accuracy over encoding speed, especially for applications where near loss-less representation is essential. We demonstrate the approach using the Cu$_3$Au sample. We begin by training a large WIEN-INR (25.23MiB), then apply post-training compression: (i) global $\ell_1$ unstructured pruning to induce sparsity, (ii) weight quantization via k-means (weight sharing), and (iii) Huffman coding of the index maps. By varying the pruning percentage (fraction of weights set to zero) and the clustering size (number of quantization centroids), we control the degree of model compression. We report the resulting model size alongside PSNR. This flexibility allows practitioners to operate WIEN-INR in a lightweight setting for interactive throughput (as shown in the main text), or in a near-lossless setting (longer training, larger model) for archival analysis.

\begin{table}[h]
\centering
\small
\begin{tabular}{r r r r r r}
\toprule
$s$ & $k$ & Size (MiB) & Comp. rate & PSNR (dB) & $\Delta$PSNR (dB) \\
\midrule
0.0 &  320 & 7.00 & 3.60 & 37.77 & $-12.15$ \\
0.0 &  640 & 7.85 & 3.21 & 37.84 & $-12.07$ \\
0.0 & 1280 & 8.70 & 2.90 & \textbf{49.85} & \textbf{$-0.06$} \\
0.0 & 2560 & 9.57 & 2.64 & \textbf{49.89} & \textbf{$-0.02$} \\
\midrule
0.1 &  320 & 6.78 & 3.72 & 37.75 & $-12.16$ \\
0.1 &  640 & 7.56 & 3.34 & 33.89 & $-16.02$ \\
0.1 & 1280 & 8.35 & 3.02 & 37.77 & $-12.14$ \\
0.1 & 2560 & 9.16 & 2.76 & 47.45 & $-2.46$ \\
\midrule
0.2 &  320 & 6.42 & 3.93 & 37.57 & $-12.35$ \\
0.2 &  640 & 7.13 & 3.54 & 37.58 & $-12.33$ \\
0.2 & 1280 & 7.86 & 3.21 & 37.60 & $-12.31$ \\
0.2 & 2560 & 8.60 & 2.94 & 41.68 & $-8.23$ \\
\bottomrule
\end{tabular}
\caption{\textbf{Quantization of INR weights.} PSNR is measured after rebuilding from (codebook, indices). “Comp.~$\times$” is relative to dense fp16 parameter storage before post-processing model compression. Note minimal PSNR drop (near-lossless) at $s{=}0$ occurs for $k\ge1280$. Original 41.54MiB at fp16.}
\end{table}

\newpage
\section{Technical}
\subsection{Discrete Wavelet Transformations} \label{appendix:wavelet}

We assume a separable wavelet orthonormal basis of $L^2(\mathbb{R}^p)$. Without loss of generality to higher-dimensional input spaces, as an example, we consider a two-dimensional
% \footnote{this generalizes to higher-dimensional input spaces.} 
discrete signal \( I[m,n] \in \mathbb{R}^d \) of spatial size \( M \times N \), where each spatial location encodes a \( d \)-dimensional vector. The multi-level two-dimensional discrete wavelet transform (DWT) is applied independently to each channel: 
% \footnote{we will drop the channel dependence $c$ in the following discussions}:
% \[
% \text{DWT}_J(I) = \left\{ \text{DWT}_J(I^{(1)}), \text{DWT}_J(I^{(2)}), \dots, \text{DWT}_J(I^{(d)}) \right\},
% \]
\begin{equation}
\begin{split}
\text{DWT}_J(I) = \bigl\{ \text{DWT}_J(I^{(1)}), \text{DWT}_J(I^{(2)}) & , \\
 \dots, \text{DWT}_J(I^{(d)}) & \bigr\},
\end{split}
\end{equation}
where \( I^{(c)}[m,n] \in \mathbb{R} \) denotes the scalar-valued channel \( c \in \{1, \dots, d\} \). We will drop the channel dependence $c$ in the following discussions for brevity.

Subsequently, each channel is decomposed into a multi-resolution representation:
\begin{equation}\label{eq:DWT}
\text{DWT}_J(I^{(c)}) = \left( \mathbf{a}_J^{(c)}, \left\{ \mathbf{d}_i^{1,(c)}, \mathbf{d}_i^{2,(c)}, \mathbf{d}_i^{3,(c)} \right\}_{i=1}^J \right),
\end{equation}
where $J$ is the number of decomposition levels. $\text{a}_J^{(c)}$ is the approximation coefficients that describe the data at the coarsest level, of size $\frac{M}{2^J} \times \frac{N}{2^J}$, while $\text{b}_i^{j,(c)}$ denote the detail coefficients at level $i$ of orientation $j$, each of size $\frac{M}{2^i} \times \frac{N}{2^i}$. Increasing $i$ corresponds to progressively coarser detail. 

This framework extends directly to higher-dimensional input spaces. Specifically, let $b[k]$ be a $p$-dimensional input sampled at intervals $2^{-L}$, associated with a function $f$ at scale $2^L$. We associate $b[k]$ to an approximation to $f$ at the scale $2^L$ with scaling coefficients $\mathbf{a}_L[k] = \langle f, \phi^0_{L,k} \rangle$ that satisfy $b[k] = 2^{-Lp/2}a_L[k]\approx f(2^L k)$, where $\phi^0_{L,k}$ is the scaling function at scale $2^L$. The wavelet coefficients at scales $2^j > 2^L$ are computed as $a_j[k] = \langle f, \psi^0_{j,k} \rangle$ and $d^\epsilon_j[k] = \langle f, \psi^\epsilon_{j,k} \rangle$ for $0 < \epsilon < 2^p$, where $\psi$ is be the corresponding wavelet generating a wavelet orthonormal basis of $L^2(\mathbb{R}^p)$. 

Generally, each successive scale approximately halves the bandwidth in the frequency domain, with the wavelet at scale $2^j,j\geq L$ approximately responding to frequencies in the octave band $[\pi/2^{j-L+1}, \pi/2^{j-L}]$, subject to some overlap between adjacent scales due to the non-ideal frequency localization of practical wavelets. We drop the $L$ in the main-text when it is clear from the context.

The coefficients can be computed using the Fast Wavelet Transform (FWT) by consecutive convolution and downsampling: $a_{j+1}[k] = [a_j * \bar{h}^0](2k)$ and $d^\epsilon_{j+1}[k] = [a^j * \bar{g}^\epsilon](2k), j>L$, where the low-pass and high-pass filters $(h, g)$ are the conjugate mirror filters associated to the wavelet $\psi$. FWT requires on the order of $d O(N \log2 N), N = N_1\times\cdots\times N_p$ operations for a signal of total length $N$. 

The choice of wavelet $\psi$ (defining $h[k]$ and $g[k]$) significantly impacts the transform's characteristics, with families like Haar, Daubechies, and Symlets offering trade-offs between compact support, smoothness, and vanishing moments. This process is invertible/lossless when $h$ and $g$ satisfy the conjugate mirror conditions, allowing exact signal reconstruction using the inverse discrete wavelet transform (IDWT)\cite{Mallat2008}. %(p349)%
\subsection{Candidate Transformations}\label{appendix:choice}
The rate-distortion performance depends on wavelet choices. While a comprehensive review of optimal wavelet transform candidates is beyond the scope of this methodology paper, we include several alternatives to orthonormal DWT that are well-established in scientific data processing. These choices can be seamlessly integrated into both the WAVELET-INR and WIEN-INR frameworks; implementation simply requires applying the appropriate forward and backward transforms, and providing suitable coordinates dimensions for the different frequency bands.

\begin{itemize}
    \item The isotropic ``\`a trous": The isotropic “à trous” (undecimated) wavelet transform provides a redundant, translation-invariant representation of signals. It is especially effective for identifying isotropic structures, and it has a long record of being used in astronomical imaging tasks where the data often contain mostly (quasi-)isotropic objects, such as star- or galaxy-like features \cite{Starck2015}. 
    Unlike the standard wavelet transform, which only decomposes the low-frequency branch at each level, the wavelet packet transform recursively decomposes both low- and high-frequency components. This produces a complete binary tree of subbands, where each node corresponds to a different time–frequency tiling. An entropy-based cost function can be applied to this tree to adaptively select the optimal basis that best matches the statistical structure of the data.
    
    \item  Wavelet Packet: Unlike the DWT that only decomposes the low-frequency branch at each level, the wavelet packet transform recursively decomposes both low- and high-frequency components and generate a complete binary tree of subbands. An entropy-based cost (best-basis) can then be applied to select the basis that best matches the statistics of a given dataset, thus a more data-driven selection of wavelet choice\cite{Coifman1993}. 
    \item The Discrete Cosine Transform (DCT) is effective for texture representation, with localized DCT variants preferred for spatially inhomogeneous textures. Ridgelets are advantageous for line-like features of fixed orientation, while curvelets provide a natural representation for edges and curved structures. In practice, DCT and curvelet dictionaries are often combined when images contain prominent edge structures \cite{Mallat2008,Starck2015}. When images exhibit geometric self-similarity, patch-based preprocessing methods can be employed to exploit local redundancies. Such strategies have also been integrated into standards like JPEG 2000 \cite{Taubman2013}.

\end{itemize}

\subsection{Definitions}\label{A:definitions}

\begin{definition}[NTK \citep{jacot2018ntk,lee2019wide}]
\label{def:lazy-ntk}
Let $\{f^{(n)}_{\theta}:\Omega\to\mathbb{R}^m\}_{n\in\mathbb{N}}$ be a depth-$L$ network family
with all hidden widths of order $n$, trained by gradient flow on squared loss over an input
distribution $\Pi$ (population) or a finite sample $\{(x_i,y_i)\}_{i=1}^N$ (empirical).
Define the Neural Tangent Kernel (NTK) at parameters $\theta$:
\[
K^{(n)}_{\theta}(x,x')\;:=\;\nabla_{\theta} f^{(n)}_{\theta}(x)\;\nabla_{\theta} f^{(n)}_{\theta}(x')^{\!\top}
\in\mathbb{R}^{m\times m}.
\]
Assume \emph{NTK parameterization} at initialization $\theta^{(n)}_0$ so that
$K^{(n)}_{\theta^{(n)}_0}$ has a non-degenerate limit as $n\to\infty$ (e.g., layer weights
$W\sim\mathcal{N}(0,\sigma^2/n)$ and $O(1)$ learning rates).
We say training is in the \emph{lazy/NTK regime on $[0,T]$} if, as $n\to\infty$,
\[
\sup_{t\in[0,T]}\big\|\theta^{(n)}(t)-\theta^{(n)}_0\big\|\;\xrightarrow{\ \mathbb{P}\ }\;0,
\qquad
\sup_{t\in[0,T]}\big\|K^{(n)}_{\theta^{(n)}(t)}-K^{(n)}_{\theta^{(n)}_0}\big\|_{\mathrm{op}}
\;\xrightarrow{\ \mathbb{P}\ }\;0,
\]
and $K^{(n)}_{\theta^{(n)}_0}\Rightarrow K_\infty$ pointwise on $\Omega\times\Omega$ for a deterministic
PSD kernel $K_\infty$. Consequently $f^{(n)}$ linearizes around $\theta^{(n)}_0$ and the predictions follow
\emph{kernel gradient flow} with a fixed kernel:
\begin{align*}
&\text{(population)}\quad
\partial_t f_t(x)\;=\;-\int_\Omega K_\infty(x,x')\big(f_t(x')-y(x')\big)\,d\Pi(x'),\quad f_{0}=f_{\theta_0},\\
&\text{(empirical)}\quad
\partial_t \mathbf{f}_t\;=\;-\Theta\,(\mathbf{f}_t-\mathbf{y}),\quad
\mathbf{f}_t=e^{-t\Theta}\mathbf{f}_0+\big(I-e^{-t\Theta}\big)\mathbf{y},
\end{align*}
where $\Theta\in\mathbb{R}^{N\times N}$ is the (block) NTK Gram matrix on $\{x_i\}$.
Equivalently,
$f^{(n)}_{\theta^{(n)}(t)}(x)=f^{(n)}_{\theta^{(n)}_0}(x)+\nabla_\theta f^{(n)}_{\theta^{(n)}_0}(x)\,
(\theta^{(n)}(t)-\theta^{(n)}_0)+o_{\mathbb{P}}(1)$ uniformly on $[0,T]$.
\end{definition}

\begin{definition}[Wavelet bands and orthogonal projectors \citep{mallat2009wavelet}]
\label{def:wavelet-bands}
Let $\Omega=[0,1]^p$ and let $\{\phi_{L,k}^0\}_k$ be scaling functions at a coarse level $L$
yielding a multiresolution analysis (MRA)
\[
V_{L}\subset V_{L+1}\subset\cdots\subset L^2(\Omega),
\qquad V_{j+1}=V_j\oplus W_j.
\]
For each scale $j$, the detail space $W_j$ is spanned by orthonormal $p$-D wavelets
$\{\psi_{j,k}^\epsilon\}_{\epsilon=1}^{2^p-1,\;k\in\mathbb{Z}^p}$ with
$\langle \psi_{j,k}^\epsilon,\psi_{j',k'}^{\epsilon'}\rangle=\delta_{jj'}\delta_{\epsilon\epsilon'}\delta_{kk'}$.
Define the \emph{wavelet band} (detail subspace) and the low-pass space by
\[
\cH_j:=W_j\quad (j\ge L),
\qquad
\cH_{\mathrm{LP}}:=V_{L},
\]
so that $L^2(\Omega)=\cH_{\mathrm{LP}}\oplus\bigoplus_{j=J_0}^{\infty}\cH_j$ is an orthogonal decomposition.
The orthogonal projector $P_j:L^2(\Omega)\to\cH_j$ acts via wavelet coefficients:
\[
(P_j f)(x)=\sum_{\epsilon=1}^{2^p-1}\sum_{k}
\big\langle f,\psi_{j,k}^\epsilon\big\rangle\,\psi_{j,k}^\epsilon(x),
\qquad
(P_{\mathrm{LP}} f)(x)=\sum_k \big\langle f,\phi_{L,k}^0\big\rangle\,\phi_{L,k}^0(x).
\]
Orthogonality implies $P_jP_{j'}=\delta_{jj'}P_j$ and
$\sum_{j=\mathrm{LP}}^{J}P_j$ is the projector onto
$\cH_{\mathrm{LP}}\oplus\bigoplus_{j=J_0}^J\cH_j$.
(On finite grids one uses a discrete wavelet transform with finite index sets; the formulas above then
hold verbatim with finite sums.)
\end{definition}

\subsection{Proofs}

\subsubsection{Proof of Lemma~\ref{lem:additive-block}}\label{a:proof_lem1}

\begin{proof}
\paragraph{Setup and notation.}
Let $(\Omega,\Pi)$ be the input space with sampling measure $\Pi$ and let
$\mathsf H:=L^2(\Omega;\mathbb R^m)$ be the output Hilbert space with inner product
$\langle u,v\rangle_{\mathsf H}=\int_\Omega u(x)^\top v(x)\,d\Pi(x)$.
For $j=0,\dots,J$, let the parameter block be a finite-dimensional Euclidean space
$\mathsf E_j\simeq\mathbb R^{p_j}$, and write the full parameter space as the orthogonal
direct sum $\mathsf E:=\bigoplus_{j=0}^J \mathsf E_j$ with the standard inner product.
Assume $f=\sum_{j=0}^J f^{(j)}$ with $f^{(j)}:\mathsf E_j\to \mathsf H$ Fr\'echet differentiable.

Fix a reference parameter $\theta=(\theta_0,\dots,\theta_J)$ and write the (Fr\'echet) Jacobian
of $f^{(j)}$ at $\theta_j$ as a bounded linear operator
\[
J_j \;\equiv\; D f^{(j)}(\theta_j) \;:\; \mathsf E_j \longrightarrow \mathsf H,\qquad
(J_j v_j)(x)\;=\;\sum_{a=1}^{p_j} v_{j,a}\,\partial_{\theta_{j,a}} f^{(j)}(\theta_j)(x).
\]
Let $J:\mathsf E\to\mathsf H$ be the block-concatenated Jacobian,
$(Jv)(\cdot):=\sum_{j=0}^J J_j v_j$, so that $J=[\,J_0\;\cdots\;J_J\,]$.
We denote by $J_j^\ast:\mathsf H\to\mathsf E_j$ and $J^\ast:\mathsf H\to\mathsf E$ their Hilbert adjoints,
defined by $\langle J_j v_j,h\rangle_{\mathsf H}=\langle v_j,J_j^\ast h\rangle_{\mathsf E_j}$ and
$\langle J v,h\rangle_{\mathsf H}=\langle v,J^\ast h\rangle_{\mathsf E}$, respectively.

In the (population) NTK/lazy regime, the Neural Tangent Kernel (operator) at $\theta$ is the
self-adjoint, positive semidefinite operator $K:\mathsf H\to\mathsf H$ given by the Jacobian Gram map
\[
K \;=\; J\,J^\ast,
\quad \text{ i.e. }\quad
(K h)(\cdot) \;=\; J\big(J^\ast h\big)(\cdot),
\qquad h\in\mathsf H.
\]
(Equivalently, in coordinates: $K(x,x')=\nabla_\theta f(\theta)(x)\,\nabla_\theta f(\theta)(x')^\top$,
and $Kh=\int_\Omega K(\cdot,x')h(x')\,d\Pi(x')$.) Define $K_j:=J_j J_j^\ast$ for each block.

\medskip
\noindent\textbf{Additivity for disjoint parameter blocks.}
Since $\mathsf E=\bigoplus_j \mathsf E_j$ and $J=[\,J_0\;\cdots\;J_J\,]$, we have
\[
J\,J^\ast \;=\; \Big(\sum_{j=0}^J J_j\Big)\Big(\sum_{k=0}^J J_k^\ast\Big)
\;=\; \sum_{j=0}^J \sum_{k=0}^J J_j J_k^\ast.
\]
But $J_j J_k^\ast:\mathsf H\to\mathsf H$ factors through $\mathsf E_j \cap \mathsf E_k$ at the level
of adjoints: for $j\neq k$, the ranges of $J_k^\ast$ lie in $\mathsf E_k$ and are orthogonal to $\mathsf E_j$,
so $J_j J_k^\ast\equiv 0$. Hence only the diagonal terms remain,
\[
K \;=\; J J^\ast \;=\; \sum_{j=0}^J J_j J_j^\ast \;=\; \sum_{j=0}^J K_j,
\]
which is exactly the asserted NTK additivity.

\medskip
\noindent\textbf{Block-diagonality under per-band training.}
Let $\{\cH_j\}_{j=0}^J$ be pairwise orthogonal closed subspaces of $\mathsf H$ with orthogonal projectors
$\{P_j\}$ (the wavelet bands), so that $\mathsf H=\bigoplus_{j=0}^J \cH_j$.
Assume that each subnetwork $f^{(j)}$ is trained \emph{only} on the band-$j$ target, i.e.,
the loss for $\theta_j$ depends on $P_j f^{(j)}$ but not on $P_{j'} f^{(j)}$ for $j'\neq j$.
Equivalently (and without loss of generality for the NTK), we can regard the effective output of
block $j$ as $P_j f^{(j)}$ so that its Jacobian is $P_j J_j$ and therefore $\mathrm{Ran}(J_j)\subseteq \cH_j$.

With $\mathrm{Ran}(J_j)\subseteq \cH_j$ we obtain, for any $h\in\mathsf H$ and any $j\neq k$,
\[
\langle h,\,J_j v_j\rangle_{\mathsf H}
\;=\; \langle P_j h,\,J_j v_j\rangle_{\mathsf H}
\quad\Rightarrow\quad
J_j^\ast h \;=\; J_j^\ast (P_j h),
\]
so $J_j^\ast$ \emph{annihilates} components of $h$ outside $\cH_j$.
Consequently, for $h\in\cH_\ell$ we have $J_j^\ast h=0$ whenever $j\neq \ell$, and thus
\[
K h \;=\; \sum_{j=0}^J J_j J_j^\ast h
\;=\; J_\ell J_\ell^\ast h \;\in\; \mathrm{Ran}(J_\ell)\;\subseteq\;\cH_\ell.
\]
Hence each band $\cH_\ell$ is an invariant subspace of $K$, and $K$ reduces on the orthogonal
decomposition $\mathsf H=\bigoplus_{j=0}^J \cH_j$ as
\[
K \;=\; \bigoplus_{j=0}^J \big(J_j J_j^\ast\big)
\;=\; \bigoplus_{j=0}^J K_j,
\]
i.e., $K$ is block-diagonal with respect to the band decomposition.

\medskip
\noindent\textbf{Remarks.}
(i) The same proof works in the empirical setting by replacing $L^2(\Omega,\Pi)$ inner products with finite
sums over samples (the Gram matrix form of the NTK). (ii) If the architecture does not enforce
$\mathrm{Ran}(J_j)\subseteq \cH_j$ exactly, it suffices that $P_{j'} J_j$ is small for $j'\neq j$; then the
off-diagonal blocks are small and $K$ is approximately block-diagonal to the same order. This completes the proof.
\end{proof}

\subsubsection{Proof of Theorem \ref{thm:block-opt}}\label{a:proof_thm_block-opt}

\begin{proof}
Let $\Phi(K):=\sum_{(j,i)\in\mathsf{HF}} P_{j,i} K P_{j,i}$ be the pinching onto HF band blocks. For $f(x)=e^{-2tx}$, it follows from operator Jensen's inequality under completely positive unital maps (pinching) that $f(\Phi(K))\le \Phi(f(K))$. For $v_{j,i}:=P_{j,i}r_0$,
\[
\|e^{-t\,\Phi(K)}v_{j,i}\|^2=\langle v_{j,i}, e^{-2t\,\Phi(K)} v_{j,i}\rangle
\ \le\ \langle v_{j,i}, \Phi(e^{-2tK}) v_{j,i}\rangle
=\langle v_{j,i}, e^{-2tK} v_{j,i}\rangle
=\|e^{-tK}v_{j,i}\|^2.
\]
Summing over $(j,i)\in\mathsf{HF}$ shows the HF objective cannot increase under pinching, hence an optimizer can be taken block–diagonal on HF. Feasibility of any block–diagonal $K_{\mathrm{blk}}=\bigoplus K_{j,i}$ with $\sum p_{j,i}\le P$ follows from Lemma~\ref{lem:additive-block} and the definition of $\mathcal{K}_{j,i}(p_{j,i})$, establishing equality of the minima.
\end{proof}

\subsubsection{Proof of Main Theorem \ref{thm:sep-leq-mono}}\label{a:proof_thm_sep-leq-mono}

\begin{proof}
By Theorem~\ref{thm:block-opt}, the HF–optimal NTK under any fixed budget is block–diagonal over HF bands and realizable by a band–split INR; hence the split design attains the global optimum among \emph{all} INRs. No monolithic design can outperform the optimum, so $P_{\mathrm{opt}}=P_{\textsf{sep}}\le P_{\textsf{mono}}$. The equality case is exactly when the monolithic optimum implements an equivalent block–diagonal kernel with the same bandwise budgets.
\end{proof}

\clearpage
\section{Additional Figures and Tables}

\begin{figure*}[hbt]
    \includegraphics[width=0.8\textwidth]{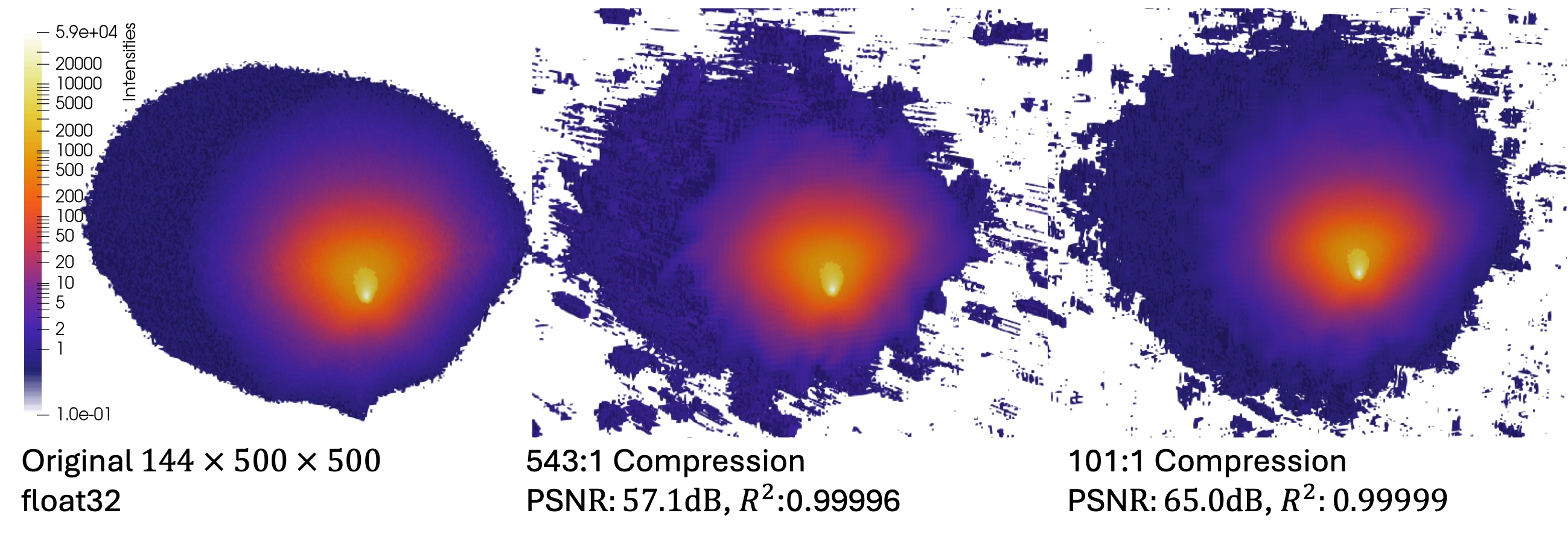}
    \caption{\textbf{A higher compression rate example on a Cu$_3$Au measurement of a different temperature \cite{ni2025cu3au}}. Our approach represents data as a neural network that maps coordinates to corresponding field values. We obtain highly compressive representations in this manner, here showing two levels of compression (middle, right) for the Cu$_3$Au sample (left). Even for extreme compression ratios (middle, 543 : 1), our method preserves high-frequency and fine details with exceptional fidelity ($R^2 = 0.99996$)}
\end{figure*}

\begin{table}[h!]
\begin{tabular}{lcrcrc}
\toprule
Method & PSNR (dB) & Hidden Features & Hidden Layers & Time (s) & Model Size (MB) \\
\midrule
FE & 36.56 & 52 & 3 & 13.129 & 0.02 \\
FE & 36.63 & 77 & 3 & 16.596 & 0.03 \\
FE & 36.67 & 113 & 3 & 20.139 & 0.06 \\
FE & 36.70 & 164 & 3 & 32.980 & 0.12 \\
FE & 36.72 & 234 & 3 & 44.382 & 0.23 \\
FE & 36.72 & 324 & 3 & 74.713 & 0.44 \\
FE & 36.72 & 465 & 3 & 121.941 & 0.88 \\
FE & 36.73 & 571 & 3 & 174.876 & 1.31 \\
FE & 36.74 & 622 & 3 & 201.026 & 1.54 \\
FE & 36.75 & 1075 & 3 & 697.670 & 4.53 \\
MLP & 36.54 & 52 & 3 & 13.203 & 0.01 \\
MLP & 36.53 & 77 & 3 & 16.781 & 0.03 \\
MLP & 36.60 & 113 & 3 & 19.765 & 0.06 \\
MLP & 36.64 & 164 & 3 & 32.062 & 0.12 \\
MLP & 36.66 & 234 & 3 & 44.271 & 0.23 \\
MLP & 36.66 & 324 & 3 & 75.138 & 0.43 \\
MLP & 36.69 & 465 & 3 & 121.907 & 0.86 \\
MLP & 36.63 & 571 & 3 & 174.844 & 1.29 \\
MLP & 36.63 & 622 & 3 & 200.922 & 1.52 \\
MLP & 36.63 & 1075 & 3 & 700.832 & 4.49 \\
SIREN & \textbf{36.69} & 52 & 3 & 18.699 & 0.02 \\
SIREN & \textbf{36.71} & 77 & 3 & 24.809 & 0.04 \\
SIREN & 36.72 & 113 & 3 & 30.872 & 0.08 \\
SIREN & 36.73 & 164 & 3 & 51.088 & 0.17 \\
SIREN & 36.75 & 234 & 3 & 72.448 & 0.33 \\
SIREN & 36.80 & 324 & 3 & 118.439 & 0.63 \\
SIREN & \textbf{37.39} & 465 & 3 & 195.332 & 1.27 \\
SIREN & 37.15 & 571 & 3 & 277.478 & 1.91 \\
SIREN & \textbf{39.15} & 622 & 3 & 317.532 & 2.26 \\
SIREN & \textbf{40.74} & 1075 & 3 & 1079.707 & 6.70 \\
TCNN & 25.26 & 52 & 3 & 32.995 & 0.04 \\
TCNN & 32.37 & 77 & 3 & 31.211 & 0.04 \\
TCNN & \textbf{36.81} & 113 & 3 & 31.577 & 0.10 \\
TCNN & \textbf{36.89} & 164 & 3 & 30.762 & 0.15 \\
TCNN & \textbf{36.99} & 234 & 3 & 30.961 & 0.25 \\
TCNN & \textbf{37.06} & 324 & 3 & 31.506 & 0.45 \\
TCNN & 37.12 & 465 & 3 & 32.739 & 0.81 \\
TCNN & \textbf{37.32} & 571 & 3 & 33.392 & 2.74 \\
TCNN & 37.42 & 622 & 3 & 31.816 & 9.07 \\
TCNN & 37.45 & 1075 & 3 & 34.776 & 16.35 \\
WIRE & 36.42 & 52 & 3 & 37.965 & 0.02 \\
WIRE & 36.60 & 77 & 3 & 60.460 & 0.04 \\
WIRE & 36.62 & 113 & 3 & 79.515 & 0.08 \\
WIRE & 36.67 & 164 & 3 & 134.582 & 0.17 \\
WIRE & 36.67 & 234 & 3 & 233.297 & 0.33 \\
WIRE & 36.70 & 324 & 3 & 360.020 & 0.63 \\
WIRE & 36.68 & 465 & 3 & 571.743 & 1.27 \\
WIRE & 36.65 & 571 & 3 & 769.351 & 1.91 \\
WIRE & 36.77 & 622 & 3 & 892.663 & 2.27 \\
WIRE & 36.76 & 1075 & 3 & 2080.022 & 6.69 \\
\bottomrule
\end{tabular}
\caption{Network configurations of the benchmarked standard INRs in representing the wide-angle X-ray scattering data. PSNR values are measured in decibels (higher is better). Bold values indicate best performance for comparable model sizes. ``Hidden Features" refers to the network width, and ``Hidden Layers" indicates the network depth for each model. All models were trained for 4000 steps.}
\label{tab:standard_methods}
\end{table}

\begin{table}[h!]
\begin{tabular}{lrcccrcr}
\toprule
Method & PSNR (dB) & Low-Freq Network & High-Freq Network & Kernel Network & Hidden Layers & Time (s) & Model Size (MB) \\
\midrule
WIEN-INR & \textbf{37.45} & 22 & 24, 24, 24 & 24 & 3 & 28.712 & 0.16 \\
WIEN-INR & \textbf{37.60} & 22 & 24, 24, 27 & 24 & 3 & 30.822 & 0.16 \\
WIEN-INR & \textbf{38.16} & 22 & 24, 24, 37 & 30 & 3 & 31.315 & 0.18 \\
WIEN-INR & \textbf{39.90} & 23 & 24, 27, 56 & 56 & 3 & 34.045 & 0.26 \\
WIEN-INR & \textbf{41.66} & 24 & 24, 37, 86 & 101 & 3 & 40.998 & 0.42 \\
WIEN-INR & \textbf{42.62} & 25 & 26, 54, 125 & 165 & 3 & 51.479 & 0.70 \\
WIEN-INR & \textbf{43.57} & 29 & 36, 83, 185 & 272 & 3 & 71.390 & 1.33 \\
WIEN-INR & \textbf{44.28} & 34 & 48, 112, 242 & 386 & 3 & 99.802 & 2.17 \\
WIEN-INR & \textbf{44.69} & 39 & 59, 136, 290 & 477 & 3 & 118.746 & 2.99 \\
WIEN-INR & 37.85 & 57 & 95, 209, 439 & 702 & 3 & 202.599 & 5.78 \\
Wavelet-INR & 36.99 & 22 & 24, 24, 24, 37 & - & 3 & 14.748 & 0.14 \\
Wavelet-INR & 37.25 & 22 & 24, 24, 27, 56 & - & 3 & 14.600 & 0.15 \\
Wavelet-INR & 37.80 & 22 & 24, 24, 37, 86 & - & 3 & 15.887 & 0.19 \\
Wavelet-INR & 38.95 & 23 & 24, 27, 56, 129 & - & 3 & 17.479 & 0.27 \\
Wavelet-INR & 40.56 & 24 & 24, 37, 86, 191 & - & 3 & 19.749 & 0.43 \\
Wavelet-INR & 41.45 & 25 & 26, 54, 125, 267 & - & 3 & 23.414 & 0.73 \\
Wavelet-INR & 42.33 & 29 & 36, 83, 185, 390 & - & 3 & 32.743 & 1.36 \\
Wavelet-INR & 42.77 & 34 & 48, 112, 242, 504 & - & 3 & 41.327 & 2.17 \\
Wavelet-INR & 43.09 & 39 & 59, 136, 290, 574 & - & 3 & 51.834 & 3.05 \\
Wavelet-INR & \textbf{43.93} & 57 & 95, 209, 439, 726 & - & 3 & 82.612 & 6.36 \\
\bottomrule
\end{tabular}
\caption{Network configurations of the proposed methods in representing the wide-angle X-ray scattering data. All subnetworks use a depth of 3 hidden layers. The ``Low-freq Network" column indicates the width of the SIREN network for the approximation band. The ``High-freq Network" column lists the network widths for each detail band, with finer bands assigned progressively larger network widths. For WIEN-INR, the enhancement module is applied only to the finest detail band; the ``Kernel Network" column shows the width of this enhancement network, which also uses a SIREN backbone. Each subnetwork was trained for 2000 steps, except for the kernel network which was trained for 4000 steps.}
\label{tab:wavelet_methods}
\end{table}

% \begin{figure}[ht]
%     \centering
%     % First subplot
%     \subfigure[First Subplot Caption]{
%         \includegraphics[width=0.45\textwidth]{new_plot/Training_Loss_All_Cu3Au.png}
%         \label{fig:subfig1}
%     }
%     \hfill
%     % Second subplot
%     \subfigure[Second Subplot Caption]{
%         \includegraphics[width=0.45\textwidth]{new_plot/Training_Loss_WAVELET_Cu3Au.png}
%         \label{fig:subfig2}
%     }
%     \caption{Overall caption describing all three subplots.}
%     \label{fig:speckle_comp}
% \end{figure}

\begin{figure}[hbt]
    \centering
    \includegraphics[width=\linewidth]{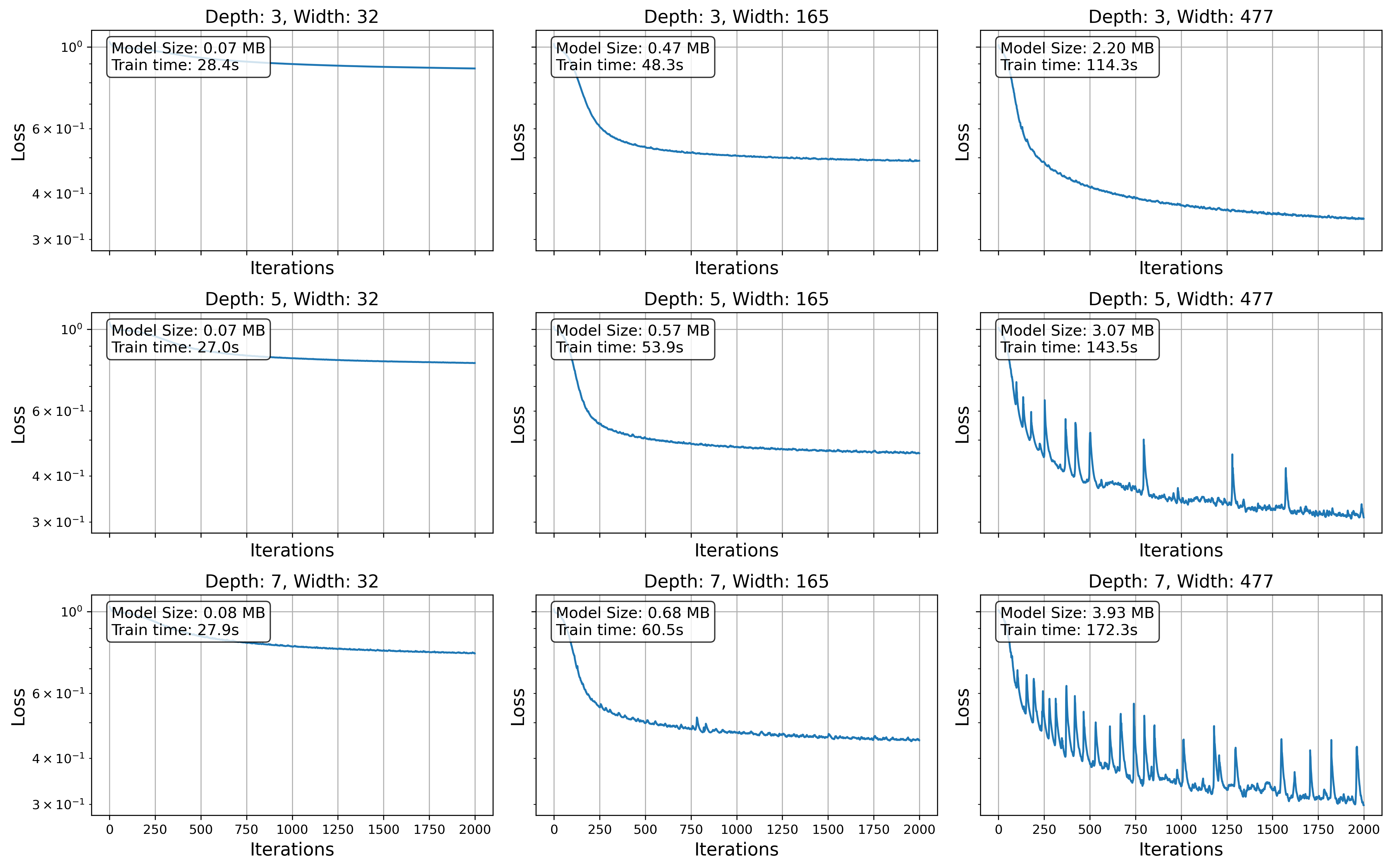}
    \caption{\textbf{WIEN-INR Convergence on Cu$_3$Au sample.} Ablation study of the convergence of the finest detail band using our enhancement module conditioned on the previous band's prediction, with varying network depth and width (learning rate is 1e-4 with weight decay 0). These curves illustrate the performance of the WIEN-INR architecture's improved capability in representing the finest-detail. Larger networks demonstrate convergence to lower loss values compared to the baseline without enhancement (Fig.~\ref{fig:convergence_Siren}), highlighting the effectiveness of the proposed enhancement scheme for high-frequency representation.
    }
    \label{fig:wien_convergence}
\end{figure}
\clearpage

\end{document}